\documentclass{article}

\usepackage[accepted]{icml2024}

\usepackage[colorlinks=true,allcolors=perfblue]{hyperref} 
\usepackage{empheq}
\usepackage[utf8]{inputenc} %
\usepackage[T1]{fontenc}    %
\usepackage{url}            %
\usepackage{booktabs}       %
\usepackage{amsfonts}       %
\usepackage{nicefrac}       %
\usepackage{microtype}      %
\usepackage{lipsum}		%
\usepackage{graphicx}
\usepackage{natbib}
\usepackage{doi}
\usepackage{amsmath}
\usepackage{algorithm}
\usepackage{algorithmic}
\usepackage{amssymb}
\usepackage{enumitem}
\usepackage{pifont}
\usepackage{tikz}
\usetikzlibrary{arrows,calc,matrix,backgrounds}
\newcommand{\tikzAngleOfLine}{\tikz@AngleOfLine}
\def\tikz@AngleOfLine(#1)(#2)#3{%
\pgfmathanglebetweenpoints{%
\pgfpointanchor{#1}{center}}{%
\pgfpointanchor{#2}{center}}
\pgfmathsetmacro{#3}{\pgfmathresult}%
}
\newcommand{\cmark}{\ding{51}}%
\newcommand{\xmark}{\ding{55}}%

\usepackage[utf8]{inputenc} %
\usepackage[T1]{fontenc}    %
\usepackage{url}            %
\usepackage{booktabs}       %
\usepackage{amsfonts, amsmath, amssymb, bm}       %
\usepackage{nicefrac}       %
\usepackage{microtype}      %
\usepackage{xcolor}         %
\usepackage{amsthm}
\usepackage{thmtools}
\usepackage{graphicx}
\usepackage{float}
\usepackage{algorithm}
\usepackage{algorithmic}
\usepackage{pythonhighlight}

\usepackage{caption}
\usepackage{subcaption}
\DeclareMathOperator*{\argmin}{\arg\!\min}
\DeclareMathOperator*{\argmax}{\arg\!\max}

\usepackage{dsfont}

\usepackage{etoc}

\usepackage[customcolors]{hf-tikz}
\definecolor{expert}{HTML}{008000}
\definecolor{error}{HTML}{f96565}
\definecolor{learner}{HTML}{F79646}
\definecolor{perfblue}{RGB}{64, 114, 175}

\theoremstyle{plain}
\newtheorem{theorem}{Theorem}[section]

\newtheorem{lemma}[theorem]{Lemma}
\newtheorem{example}[theorem]{Example}
\newtheorem{corollary}[theorem]{Corollary}
\theoremstyle{definition}
\newtheorem{definition}[theorem]{Definition}

\newtheorem{assumption}[theorem]{Assumption}
\theoremstyle{remark}

\declaretheoremstyle[
headfont=\normalfont\itshape,
qed=\qedsymbol,
]{mypf}
\declaretheorem[numbered=no, name=Proof, style=mypf]{pf}
\declaretheorem[numbered=no, name=Proof Sketch, style=mypf]{sk}

 \renewcommand{\algorithmiccomment}[1]{// #1}
 \usepackage{bbm}

\usepackage{transparent}
\newcommand{\algcommentlight}[1]{\textcolor{perfblue}{\transparent{0.8}\small{\texttt{\textbf{//\hspace{2pt}#1}}}}}

\usepackage{nicematrix}

\icmltitlerunning{A Minimaximalist Approach to Reinforcement Learning from Human Feedback}

\begin{document}

\twocolumn[
\icmltitle{A Minimaximalist Approach to \\ Reinforcement Learning from Human Feedback}

\icmlsetsymbol{equal}{*}

\begin{icmlauthorlist}
\icmlauthor{Gokul Swamy}{yyy}
\icmlauthor{Christoph Dann}{comp}
\icmlauthor{Rahul Kidambi}{comp}
\icmlauthor{Zhiwei Steven Wu}{yyy}
\icmlauthor{Alekh Agarwal}{comp}
\end{icmlauthorlist}

\icmlaffiliation{yyy}{Carnegie Mellon University}
\icmlaffiliation{comp}{Google Research}

\icmlcorrespondingauthor{Gokul Swamy}{gswamy@cmu.edu}

\icmlkeywords{Machine Learning, ICML}

\vskip 0.3in
]

\printAffiliationsAndNotice{Most of this work was completed while GS was a Student Researcher at Google Research.} %

\begin{abstract}
We present \textit{Self-Play Preference Optimization} (\texttt{SPO}), an algorithm for reinforcement learning from human feedback. Our approach is \textit{minimalist} in that it does not require training a reward model nor unstable adversarial training and is therefore rather simple to implement. Our approach is \textit{maximalist} in that it provably handles non-Markovian, intransitive, and stochastic preferences while being robust to the compounding errors that plague offline approaches to sequential prediction. To achieve the preceding qualities, we build upon the concept of a \textit{Minimax Winner} (MW), a notion of preference aggregation from the social choice theory literature that frames learning from preferences as a zero-sum game between two policies. By leveraging the symmetry of this game, we prove that rather than using the traditional technique of dueling two policies to compute the MW, we can simply have a \textit{single} agent play against itself while maintaining strong convergence guarantees. Practically, this corresponds to sampling multiple trajectories from a policy, asking a \textit{preference} or teacher model to compare them, and then using the proportion of wins as the reward for a particular trajectory. We demonstrate that on a suite of continuous control tasks, we are able to learn significantly more efficiently than reward-model based approaches while maintaining robustness to the intransitive and stochastic preferences that frequently occur in practice when aggregating human judgments. 
\end{abstract}

\section{Introduction}
\label{sec:intro}
Reinforcement learning from human feedback (RLHF, \citet{christiano2017deep}) also known as preference-based reinforcement learning (PbRL, \citet{akrour2012april, wirth2017survey, sadighactive, ibarz2018reward, lee2021b, lee2021pebble, sikchi2022ranking}), is a technique for policy optimization based on relative, rather than absolute, feedback. Owing to the relative ease of providing comparative feedback rather than absolute scores for agent behavior for human raters \citep{miller1956magical}, RLHF has been successfully applied across fields from robotics \citep{zucker2011optimization, cakmak2011human, tucker2020preference, swamy2020scaled, biyik2020active} to recommendation \citep{de2009preference, ailon2010preference, viappiani2010optimal, afsar2022reinforcement}, to retrieval \citep{yue2009interactively}. As of late, RLHF has attracted renewed interest as a leading technique for fine-tuning large language models (LLMs) \citep{ziegler2020finetuning, stiennon2020learning, bai2022training, ouyang2022training}.

\addtolength{\belowcaptionskip}{-10pt}
\begin{figure}
     \centering
     \includegraphics[width=0.4\textwidth]{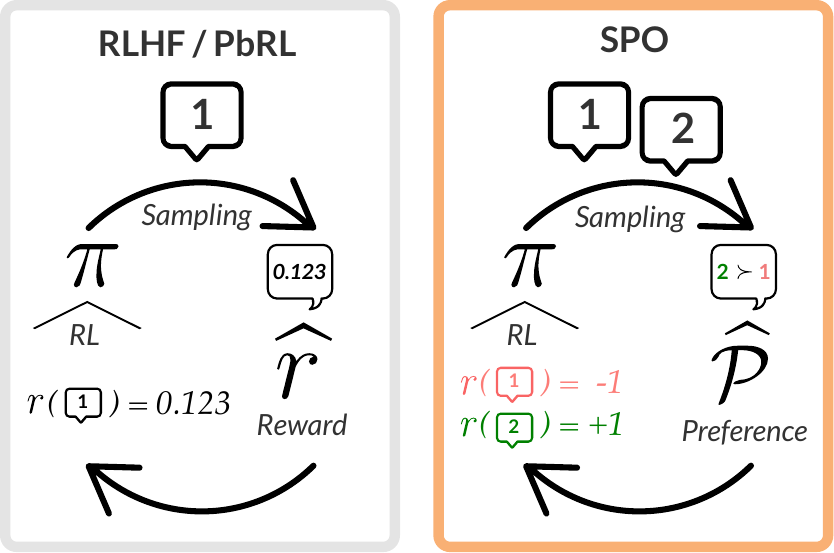}
     \caption{The standard pipeline (left) for preference-based RL / RLHF involves training a \textit{reward} model (i.e. a classifier) based on a dataset of pairwise preferences and then optimizing it via RL. We introduce \texttt{SPO} (right), a method that instead optimizes directly based on preference feedback provided by a \textit{preference} or teacher model, with each trajectory getting a reward based on the proportion of other on-policy trajectories it is preferred to. We prove and validate empirically that this approach is more robust to intransitive, non-Markovian, and noisy preferences than prior works.}
     \label{fig:ffig}
 \end{figure}
 \begin{figure*}[t]
    \centering
    \begin{subfigure}[b]{.24\linewidth}
    \centering
    \includegraphics[width=\linewidth]{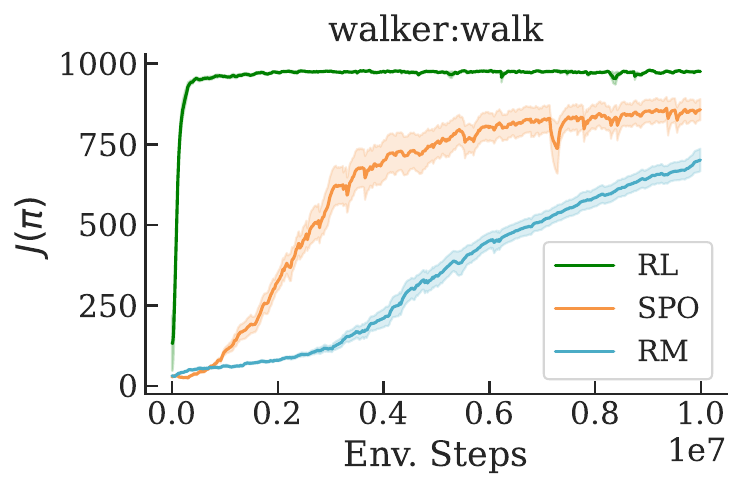}
    \caption{Transitive Preferences}
    \end{subfigure} 
    \begin{subfigure}[b]{.24\linewidth}
    \centering
    \includegraphics[width=\linewidth]{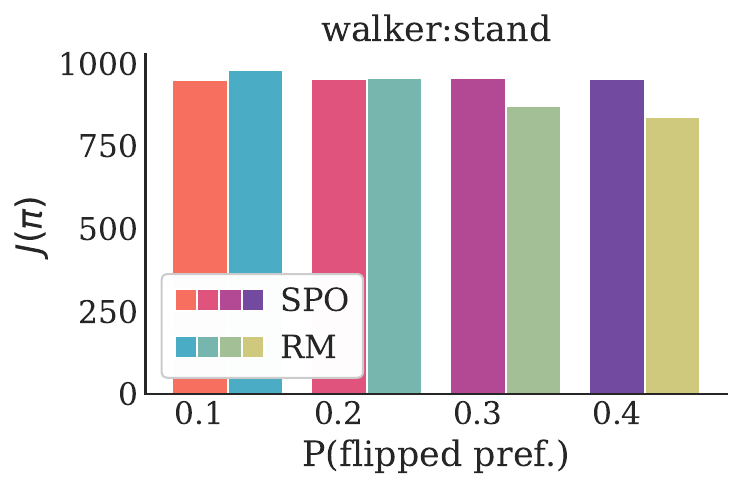}
    \caption{Stochastic Preferences}
    \end{subfigure}
    \begin{subfigure}[b]{.24\linewidth}
    \centering
    \includegraphics[width=0.6\linewidth]{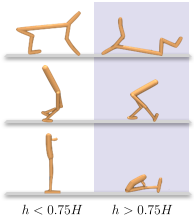}
    \caption{Non-Markovian Preferences}
    \end{subfigure}
    \begin{subfigure}[b]{.24\linewidth}
    \centering
    \includegraphics[width=0.85\linewidth]{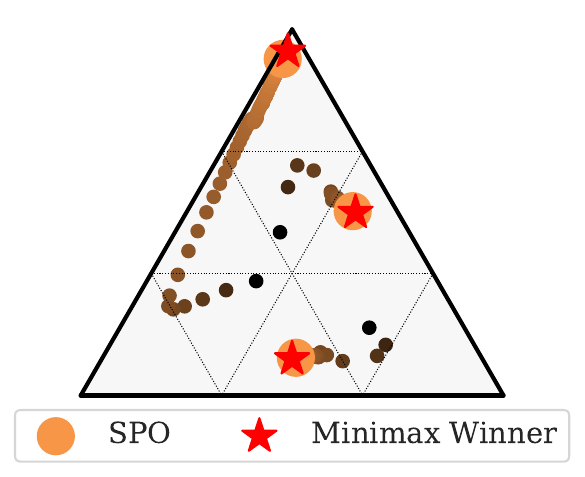}
    \caption{Intransitive Preferences}
    \end{subfigure}
    \caption{\footnotesize\texttt{SPO} gives us a unified approach to optimize a variety of preference structures. Starting with the ``nicest" case where preferences are consistent with a single reward signal \textit{\textbf{(a)}}, \texttt{SPO} is more sample-efficient than iterative Reward Modeling (RM) based approaches. Moreover, \texttt{SPO} learns comparable to RM-based approaches with stochastic preferences, where we flip preferences corresponding to a ground truth reward, without the burden of an extra model \textit{\textbf{(b)}}. Furthermore, \texttt{SPO} handles complex non-Markovian preferences \textit{\textbf{(c)}} such as learning to maximize rewards over the first three-quarters of a trajectory while not crossing a threshold on returns in the last quarter, despite searching over a class of Markovian policies. Lastly, considering intransitive preferences generated by aggregating sub-populations where a single reward function cannot explain the preferences, \texttt{SPO} computes a Minimax Winner (MW) consistently across problem instances \textit{\textbf{(d)}}. In Fig.~\ref{fig:max_reward}(a), we show that the \emph{RM based approach fails} in this setting and always converges to a deterministic policy.}
    \label{fig:teaser}
\end{figure*}
\addtolength{\belowcaptionskip}{10pt}

The predominantly studied approach to RLHF is via \textit{Reward-based RLHF}, a two-stage procedure. First, given pairs of preferred and dis-preferred behavior, one trains a \textit{reward model} to assign higher scores to the former via a classification objective. One then optimizes this reward function via some reinforcement learning algorithm. %

Simple as the above recipe is, the key ingredient of a reward model can have some undesirable effects. First, assuming an underlying reward function exists is equivalent to assuming that there exists a \textit{total order} over agent behavior. This means that there are no intransitivities in rater preferences (i.e. $A \succ B, B \succ C \Rightarrow A \succ C$), which contradicts what psychology tells us about actual human decision making \citep{tversky1969intransitivity, Gardner_2013}. Even if one believes an \textit{individual} person's preferences are transitive, when aggregated across a \textit{population} of raters, as is necessary at scale, transitivity is unlikely to be satisfied \citep{may1954intransitivity}. Second, given the inherent stochasticity of human preferences \citep{agranov2017stochastic}, one often learns a reward model that leads to a collapse in generation diversity. For example, consider a problem where the agent can pick one of two options, each of which is preferred by a sub-population of raters that makes up half of the total population. Due to either finite sample or optimization error, we can easily learn a model that assigns a slightly higher reward to one option over the other. Then, if we were to optimize our policy under this model, we would learn to (almost) exclusively select one option, leaving half of the population unsatisfied.

\begin{table*}[h]
\begin{center}
\begin{small}
\setlength{\tabcolsep}{2pt}
\begin{tabular}{>{\centering\arraybackslash}m{3.5cm}>{\centering\arraybackslash}m{4cm}>{\centering\arraybackslash}m{3cm}>{\centering\arraybackslash}m{2cm}>{\centering\arraybackslash}m{2.5cm}>{\centering\arraybackslash}m{2.5cm}>{\centering\arraybackslash}m{5cm}}
\toprule
Approach & Example & \mbox{Compounding Errors} & \mbox{Intransitive Prefs} & Learning Setup \\
\midrule
Offline, Reward-Based & DPO \citep{rafailov2023direct} &  {\color{error}\xmark}, Example \ref{ex:orbits} & {\color{error}\xmark}, Theorem \ref{thm:cyclic} & Offline, log-loss\\ 
\midrule
Online, Reward-Based & PPO \citep{ouyang2022training} & {\color{expert}\cmark}  & {\color{error}\xmark}, Theorem \ref{thm:cyclic} & Online RL \\
\midrule
Online, Dueling  & DBGD \citep{yue2009interactively}   & {\color{expert}\cmark} & {\color{expert}\cmark} & \mbox{Online \emph{adver.} RL}\\ \midrule
Online, Preference-Based  &  \textbf{\texttt{SPO}} (ours)  & {\color{expert}\cmark} & {\color{expert}\cmark} & Online RL\\
\bottomrule
\end{tabular}
\end{small}
\end{center}
\vskip -0.1in
\caption{An taxonomy of RLHF algorithms and the sorts of issues they are robust to. \label{onetabletorulethemall} }
\end{table*}

In recognition of the above concerns, various reward-model-free approaches have been proposed in the prior literature. One particularly promising set of techniques frames RLHF as a two-player zero-sum game between two policies, each of which attempts to produce behavior that is preferred by a rater to the other's \citep{yue2012k}. While elegant theoretically, this ``dueling'' framing inherits the inherent instability of adversarial training in practice and has therefore mostly been applied to bandit problems \citep{dudik2015contextual, saha2021adversarial, saha2022efficient} (with some recent exceptions: \citet{wang2023rlhf, munos2023nash}).

Motivated by these issues, we provide a simple, theoretically rigorous, and empirically performant approach to RLHF that eliminates reward modeling and does not require adversarial training. Our approach follows from two key insights. First, \textbf{\textit{by framing RLHF as a two-player zero-sum game, we are able to truly eliminate reward models}} and therefore are robust to noisy, intransitive, and non-Markovian preferences that frequently occur in practice. Second, \textbf{\textit{by leveraging the symmetry of the game, we prove that we can simply train a single agent in a self-play fashion}}, eliminating the need for unstable adversarial training. Practically, this corresponds to sampling multiple trajectories from the agent, asking a \textit{preference} or teacher model to compare each pair, and setting the reward to be the trajectory's win rate. We call our approach \textbf{\texttt{SPO}}: Self-Play Preference Optimization. More explicitly, our contributions are as follows:

\noindent \textbf{1. We derive \texttt{SPO}: an algorithm for RLHF that avoids reward modeling, compounding errors, and adversarial training.} By building upon the concept of a \textit{Minimax Winner} from social choice theory, we are able to frame RLHF as a two-player zero-sum game. We then leverage the symmetry of the payoff matrix of this game to prove that we can simply train a single agent against itself.

\noindent \textbf{2. We use a reduction-based analysis to investigate the convergence properties of \texttt{SPO}.} When intransitive preferences exist, we prove that \texttt{SPO} converges to an approximate Minimax Winner at the rate of the underlying no-regret algorithm. We also prove that when an underlying reward function does exist, our approach converges to the optimal policy at a fast rate that matches that of standard techniques.

\noindent \textbf{3. We demonstrate that on a suite of continuous control tasks with realistic preference functions, \texttt{SPO} is more performant than reward-model based approaches.} We find that our approach is able to learn more sample-efficiently than reward-model based approaches across a variety of preference setups. This includes trajectory-level comparisons based on ground-truth Markovian rewards in the easiest case, stochastic preferences, trajectory-level non-Markovian preferences and intransitive preferences induced by aggregating over sub-populations. The strong performance of \texttt{SPO} in the latter three challenging setups, all motivated by practical situations, is illustrated in Figure~\ref{fig:teaser}.

\textbf{Due to limited space, we discuss related work in detail in Appendix \ref{app:rel}.} Table~\ref{onetabletorulethemall} describes the relationships between and relative benefits of the different approaches to RLHF.

\section{Reinforcement Learning from Human Feedback via Game Solving}
We begin by introducing the notation we will use throughout the paper before defining our solution concept and deriving an efficient algorithm to compute it.

\subsection{Preliminaries} Consider a finite-horizon reward-free Markov Decision Process (MDP) \citep{puterman2014markov} parameterized by $\langle \mathcal{S}, \mathcal{A}, \mathcal{T}, H \rangle$ where $\mathcal{S}$, $\mathcal{A}$ are the state and action spaces, $\mathcal{T}: \mathcal{S} \times \mathcal{A}\rightarrow \Delta(\mathcal{S})$ is the transition operator, and $H$ is the horizon. \footnote{We omit contexts for simplicity of presentation but they can be added without much overhead, as we detail in Appendix \ref{app:proofs:context}.} We use $\Xi \triangleq (\mathcal{S} \times \mathcal{A}) ^H$ to denote the space of trajectories and $\Phi_h \triangleq \times (\mathcal{S} \times \mathcal{A}) ^{h-1} \times \mathcal{S}$ to denote the space of histories of length $h$. 

\noindent \textbf{Preference Oracle.} In the preference-based RL setup, we are given query access to a \textit{preference function}\begin{equation}
    \mathcal{P}: \Xi \times \Xi \rightarrow [-1, 1] 
\end{equation}
which, given two trajectories $\xi_1, \xi_2 \in \Xi \times \Xi$, outputs a scalar that indicates the preferred trajectory. Explicitly, given some comparison function $P(\xi_1 \succ \xi_2)$, we define $\mathcal{P}(\xi_1, \xi_2) = 2 P(\xi_1 \succ \xi_2) -1 $.
Practically, this could be either be a preference model trained on an offline dataset (RLAIF, \citet{bai2022constitutional,munos2023nash,zhao2023slic}) or a human-in-the-loop (RLHF, \citet{tucker2020preference}). The former setup is similar to reward-based RLHF, except that the reward model learning step is replaced by learning a pairwise preference model, which is more natural when learning from pairwise preference data. \footnote{e.g. $\widehat{\mathcal{P}} = \argmin_{\widetilde{\mathcal{P}}} \mathbb{E}_{(\xi^{+}, \xi^{-}) \sim \mathcal{D}}[- \log(\frac{1}{2}(\widetilde{\mathcal{P}}(\xi^{+}, \xi^{-}) + 1)]$.} Viewed this way, the preference-model based RLHF based methodology strictly generalizes reward-based RLHF, as we can still represent a preference function that is induced by a difference of rewards, but not every preference function is expressible this way, as we illustrate in the following sections. Also, in the ``alignment'' of generative models the preference function sometimes consists of a prompted generative model that is asked to compare two outputs, rather than a model trained explicitly on a preference dataset~\citep{bai2022constitutional}. In this setting, we are able to optimize directly based on the outputs of such a model, rather than requiring a reward model detour -- \textbf{see Figure \ref{fig:wrkflw} for both full workflows.}

By construction, preference functions are anti-symmetric, i.e. $\forall \xi_1, \xi_2 \in \Xi \times \Xi $, $\mathcal{P}(\xi_1, \xi_2) = -\mathcal{P}(\xi_2, \xi_1)$. Similarly, we have that $\forall \xi \in \Xi$, $ \mathcal{P}(\xi, \xi) = 0$. 

We assume access to a convex and compact policy class $\Pi \subseteq \{\mathcal{S} \rightarrow \Delta(\mathcal{A})\}$. 
With a slight abuse of notation, we can now define the preference function over policy pairs as
\begin{equation}
    \mathcal{P}(\pi_1, \pi_2) \triangleq \mathbb{E}_{\xi_1 \sim \pi_1, \xi_2 \sim \pi_2 }[\mathcal{P}(\xi_1, \xi_2)]. 
\end{equation}
\subsection{A Brief Introduction to Social Choice Theory}\label{sec:game_theory_primer}
Given choices from a population of raters that are represented as a preference function $\mathcal{P}$, social choice theory \citep{sen1986social} studies the question of how best to select options that satisfy the diversity of preferences inherent in the said population. For example, consider the set of preferences $\mathcal{P}_1$ over options $(a, b, c, d)$ in Figure \ref{fig:pm}.
\begin{figure}[!h]
\centering
  \begin{minipage}[c]{0.2\textwidth}

\begin{tikzpicture}
[every node/.style={font=\footnotesize}]
\matrix [matrix of math nodes,
         nodes={rectangle, 
                minimum size=1.75em, text depth=0.25ex,
                inner sep=0pt, outer sep=0pt,
                anchor=center},
         column sep=-0.5\pgflinewidth,
         row sep=-0.5\pgflinewidth,
         inner sep=0pt,
         ] (m)
{
  & a & b & c & d \\
a & 0 & {\color{expert}+1} & {\color{expert}+1} & {\color{error}-1}  \\
b & {\color{error}-1} & 0 & {\color{expert}+1} & {\color{error}-1} \\
c & {\color{error}-1} & {\color{error}-1} & 0 & {\color{expert}+1} \\
d & {\color{expert}+1} & {\color{expert}+1} & {\color{error}-1} & 0 \\
};
\begin{scope}[on background layer]
    \filldraw[lightgray!40, rounded corners] (m-1-2.north west) -- 
        (m-1-2.south west) -- (m-1-5.south east)-- (m-1-5.north east)-- 
        cycle;
    \filldraw[lightgray!40, rounded corners] (m-2-1.north west) -- 
        (m-2-1.north east) -- (m-5-1.south east)-- (m-5-1.south west)-- 
        cycle;
\end{scope} 
\end{tikzpicture}
\end{minipage}
\begin{minipage}[c]{0.25\textwidth}
\caption{An intransitive preference function $\mathcal{P}_1$ over $(a, b, c, d)$. $\mathcal{P}_1(x, y) = 1$ if $P(x \succ y) = 1$, $-1$ if $P(x \succ y) = 0$, and $0$ if $P(x \succ y) = 0.5$. Observe that there is no unique Copeland Winner. \label{fig:pm}}
\end{minipage}
\end{figure}

Given this preference function, perhaps the most natural idea would be to pick the option that beats the largest number of other options. In the above matrix, this would be either option $a$ or $d$ as they have the largest row sums. More formally, this technique is known as a \textit{Copeland Winner} and can be expressed mathematically as
\begin{equation}
    \mathsf{CW}(\mathcal{P}) \triangleq \argmax_{\pi \in \Pi} \sum_{\pi' \in \Pi}\mathcal{P}(\pi, \pi'). 
\end{equation}
While intuitively appealing, Copeland Winners are often not unique as in our above example, raising the question of how to break ties. For example, if half of the group feels like $a \succ d$ and the other half like $d \succ a$, picking either option would leave half of the group unsatisfied. This problem only gets worse as the number of options to choose between increases, as there is unlikely to be a single option that \textit{everyone} prefers to \textit{every} other option \citep{dudik2015contextual}.\footnote{For example, we see empirical evidence of this point in the high rates of inter-annotator disagreement \citep{taori2023stanford, touvron2023llama} in LLM finetuning datasets.}

In essence, approaches that train reward models like reward-based RLHF (or implicitly assume them like DPO) are akin to computing Copeland Winners. 
Observe that our above matrix has an \textit{intransitivity}: $a \succ c, c \succ d, d \succ a$. This means that \textit{no} reward function can explain the above preferences as it would need to satisfy $r(a) > r(c)$, $r(c) > r(d)$ and $r(d) > r(a)$ simultaneously, an impossibility. Thus, the model is forced to tie-break between $a$ and $d$, potentially leaving half of the population rather unsatisfied. In practice, this tie-breaking is performed based on the incredibly noisy data used to train the reward model \citep{taori2023stanford, touvron2023llama}, making it entirely arbitrary. When combined with the fact that an $\epsilon$ difference in reward model outputs can lead to an entirely different optimal policy, we are left with an unsatisfying solution. %

One potential solution to the issues with the Copeland Winner is to \textit{randomize}. For example, we could attempt to pick a distribution over options such that we prefer samples from this distribution to those from any other distribution with probability at least $\frac{1}{2}$. For $\mathcal{P}_1$, this would correspond to us picking $(a, c, d)$, each with probability $\frac{1}{3}$, as
\begin{equation}
    \min_{z \in \{a, b, c, d \}} (\mathcal{P}_1(a, z) + \mathcal{P}_1(c, z) + \mathcal{P}_1(d, z)) / 3= 0. \nonumber
\end{equation}
Intuitively, this means that while we don't always make everyone happy (an impossibility, \citet{arrow1950difficulty, satterthwaite1975strategy}), we never pick a solution that makes a significant portion of the population consistently unhappy. Readers familiar with game theory might recognize that the above corresponds to computing the Nash equilibrium of the two-player zero-sum (2p0s) game with payoffs given by the preference function. Formally, we define the \textit{Minimax Winner} (MW, \citet{kreweras1965aggregation, 10.2307/1880533, RePEc:ecm:emetrp:v:41:y:1973:i:2:p:285-97, fishburn1984probabilistic}), also known as a \textit{von Neumann Winner} \citep{dudik2015contextual}, as the following pair of strategies:
\begin{align}
    \mathsf{MW}(\mathcal{P}) \triangleq \big(&\argmax_{p \in \Delta(\Pi)} \min_{q  \in \Delta(\Pi)} \mathbb{E}_{\pi_1 \sim p, \pi_2 \sim q}[\mathcal{P}(\pi_1, \pi_2)], \nonumber \\
    &\argmin_{q \in \Delta(\Pi)} \max_{p  \in \Delta(\Pi)} \mathbb{E}_{\pi_1 \sim p, \pi_2 \sim q}[\mathcal{P}(\pi_1, \pi_2)]\big). \nonumber
    \label{eq:mw}
\end{align}
Via Sion's minimax theorem \citep{sion1958general}, we can guarantee that the above solution concept \textit{always} exists, unlike a unique Copeland Winner. We note that because we assumed $\Pi$ is convex, we are \textit{always} able to collapse down any distribution $p \in \Delta(\Pi)$ to a \textit{single} policy $\widetilde{p} \in \Pi$ while preserving solution quality by performing a weighted average:
\begin{equation}
     \widetilde{p}(\xi) = \sum_{\pi \in \Pi} p(\pi) p(\xi \vert \pi).
\end{equation}
In short, this means that we never need to explicitly maintain a distribution over policies in practice. %
We conclude with a few observations about MWs. First, nowhere in defining a MW did we need to assume the existence of an underlying reward function, rendering the above solution concept \textit{truly} reward model-free. Second, in the case where there actually does exist an underlying reward function that explains the observed preferences, the MW coincides with the optimal policy for that reward \citep{dudik2015contextual}, rendering the MW a strict generalization of the CW. Third, MWs satisfy a variety of desirable \textit{consistency} properties (e.g. merging populations that agree on a MW cannot change the outcome, which is especially important when attempting to re-use preference datasets), which deterministic options like the CW cannot satisfy simultaneously \citep{brandl2016consistent}.
We now turn our attention to efficiently computing MWs.

\subsection{One Player is All You Need for RLHF}
Efficient algorithms for computing Nash equilibria of 2p0s games are a central focus in computational game theory. A popular approach is to run two no-regret algorithms (e.g. Hedge \citep{freund1997decision} or Online Gradient Descent \citep{zinkevich2003online}) against each other, known more commonly as adversarial training \citep{goodfellow2014generative}. When applied to computing MWs, this is known as \textit{dueling} \citep{yue2012k} and is commonly applied in the bandit setting. While elegant in theory, this technique inherits all of the notorious instabilities of adversarial training when function approximation is introduced.
Even ignoring optimization issues, simply storing both models in memory might be difficult in our current era of ``foundation'' models \citep{bommasani2021opportunities}. These issues may explain why dueling techniques have seen limited practical use.

In light of the above difficulties, we ask a simple question: \textbf{\textit{do we actually need two players to compute MWs?}} We  prove rigorously that we only need a \textit{single} player due to the anti-symmetry of preference functions. All proofs for this section can be found in Appendix \ref{app:proofs}.

First, we prove that there \textit{always} exists a symmetric MW. 
\begin{lemma} 
$\exists (\hat{p}, \hat{q}) \in \mathsf{MW}(\mathcal{P})$ s.t. $\hat{p} = \hat{q}$. \hyperref[pf:lem:sym]{[Proof]}
\label{lem:sym}
\end{lemma}

Next, we prove that we can compute a symmetric MW by running a single no-regret algorithm against its own iterates. We assume access to the following optimization oracle.
\begin{definition}
$\mathcal{O}$ is a \textit{no-regret online linear optimization algorithm over $\Delta(\Pi)$} if it produces iterates $p_{t+1} = \mathcal{O}(\ell_{1:t}) \in \Delta(\Pi)$ such that, for any sequence of $T$ linear loss functions of the form $\ell_t(p) = \mathbb{E}_{\pi \sim p}[f_t(\pi)] \in [-1, 1]$ (with $f_t~:~\Pi\to[-1, 1]$), we have
\begin{equation}
    \sum_{t=1}^T \ell_t(p_t) - \min_{p^{\star} \in \Delta(\Pi)} \sum_{t=1}^T \ell_t(p^{\star}) \leq \mathsf{Reg}(T),
\end{equation}
with $\lim_{T \to \infty} \frac{\mathsf{Reg}(T)}{T} = 0.$ 
\label{def:nr}
\end{definition}
Common algorithms like gradient descent satisfy this property \citep{zinkevich2003online}. See \citet{hazan2016introduction} for a more extensive list. We define the \textbf{\texttt{SPO} Loss} at round $t \in [T]$ as the negative preference against the current iterate $p_t$,
\begin{equation}
    \ell_t^{\texttt{SPO}}(p) \triangleq \mathbb{E}_{\pi \sim p, \pi' \sim p_t}[-\mathcal{P}(\pi, \pi'))].
\end{equation}

We can now state our main result.
\begin{theorem}
Consider a single copy of an algorithm $\mathcal{O}$ which satisfies Definition \ref{def:nr}. Initialize $p_1 \in \Delta(\Pi)$ and set $p_{t+1} = \mathcal{O}(\ell^{\texttt{SPO}}_{1:t})$. Then, $\bar{p} = (p_1+\ldots+p_T)/T$ is a $\frac{2 \mathsf{Reg}(T)}{T}$-approximate Minimax Winner. \hyperref[pf:thm:spo-ff]{[Proof]}
\label{thm:spo-ff}
\end{theorem}
For ease of presentation, Theorem \ref{thm:spo-ff} is stated and proved assuming \textit{full feedback}, where we observe $\ell_t^{\texttt{SPO}}(p_t)$ at round $t$. In Appendix \ref{app:proofs:bandits}, we generalize the argument to the more realistic \textit{bandit feedback} setting, where we only observe a preference $\mathcal{P}(\pi, \pi')$ for some $\pi, \pi'$.

\begin{sk}
Consider a pair of strategies, $p, q \in \Delta(\Pi)$. Define $\ell_t^1(p) = \mathbb{E}_{\pi \sim p, \pi' \sim q_t}[-\mathcal{P}(\pi, \pi'))]$ and $\ell_t^2(q) = \mathbb{E}_{\pi \sim p_t, \pi' \sim q}[\mathcal{P}(\pi, \pi'))]$. By the results of \citet{freund1997decision}, we know that updating $p_{t+1} = \mathcal{O}( \ell_{1:t}^1)$ and $q_{t+1} = \mathcal{O}(\ell_{1:t}^2)$ implies that average strategies $\bar{p} = \frac{1}{t} \sum_i^t p_i$,  $\bar{q} = \frac{1}{t} \sum_i^t q_i$ converge to a Nash equilibrium (Minimax Winner) at the rate of the underlying no-regret algorithm. By construction, we  set $p_0 = q_0$. This implies that $\ell_0^1(p) =  \mathbb{E}_{\pi \sim p, \pi' \sim q_0}[-\mathcal{P}(\pi, \pi'))] =  \mathbb{E}_{\pi \sim p, \pi' \sim p_0}[-\mathcal{P}(\pi, \pi'))] = \mathbb{E}_{\pi \sim p_0, \pi' \sim p}[-\mathcal{P}(\pi, \pi'))] = \ell_0^2(p)$ due to the anti-symmetry of $\mathcal{P}$. We now proceed by induction. If, at some time $\tau$, $p_{\tau}$ and $q_{\tau}$ have the same strategy and perform updates based on the same loss function, for any deterministic $\mathcal{O}$, $p_{\tau+1} = q_{\tau+1}$. For randomized $\mathcal{O}$, our argument still applies as we describe in the full proof.
Then, by the anti-symmetry of $\mathcal{P}$, we have that $\ell_{\tau+1}^1 = \ell_{\tau+1}^2$. Hence, by induction, we have that $p_t = q_t$, $\forall t \in [T]$.%
\end{sk}

The above result implies that if we run our algorithm for long enough, we can get arbitrarily close to an exact MW.\footnote{For last iterate (rather than average iterate) convergence, one can simply set the no-regret algorithm to be Optimistic Mirror Descent and apply the results of \citet{daskalakis2017training}.}
Observe that we didn't need to assume we were running a particular algorithm $\mathcal{O}$, rendering the above a \textit{reduction} of computing minimax winners to no-regret online learning.\footnote{We note that in contrast to the usual asymptotic convergence guarantees one gets for self-play \citep{leslie2006generalised}, we inherit the rate of the underlying no-regret algorithm.} 

The above update can also be viewed from the perspective of a \textit{dynamic reward model}: it is equivalent to performing an RL step with a \textit{\textbf{policy-dependent reward model}}:
\begin{equation}
    r_{t}^{\texttt{SPO}}(\xi) \triangleq \mathbb{E}_{\substack{\pi' \sim p_t}} [\mathbb{E}_{\substack{\xi' \sim \pi'}}[\mathcal{P}(\xi, \xi')]]. \label{eq:pirm}
\end{equation}
This reward model incentivizes the learner to play trajectories that are preferred to its current distribution. Game-solving amounts to repeatedly taking a small step along this direction before (implicitly) updating the reward model. Thus, one can view \texttt{SPO} as using a perfectly shaped \textit{curriculum} to gently guide the learner.
We now pause and further contextualize our results by considering a few questions.

\noindent \textbf{Q1: \textit{Do reward-based RLHF algorithms also compute MWs?}} 
We prove that this is not the case in general by analyzing multiple algorithms which assume that there exists a reward function that explains the observed preferences, even if it is not maintained explicitly, e.g., as in Direct Preference Optimization (DPO, \citet{rafailov2023direct}).
\begin{theorem}
There exists a $\mathcal{P}$ and reference policy $\pi_{\text{ref}}$ such that the optimal policies of reward-based RLHF and DPO are not the Minimax Winner. \hyperref[pf:thm:cyclic]{[Proof]} 
\begin{figure*}[t]
\centering
    \includegraphics[width=0.8\textwidth]{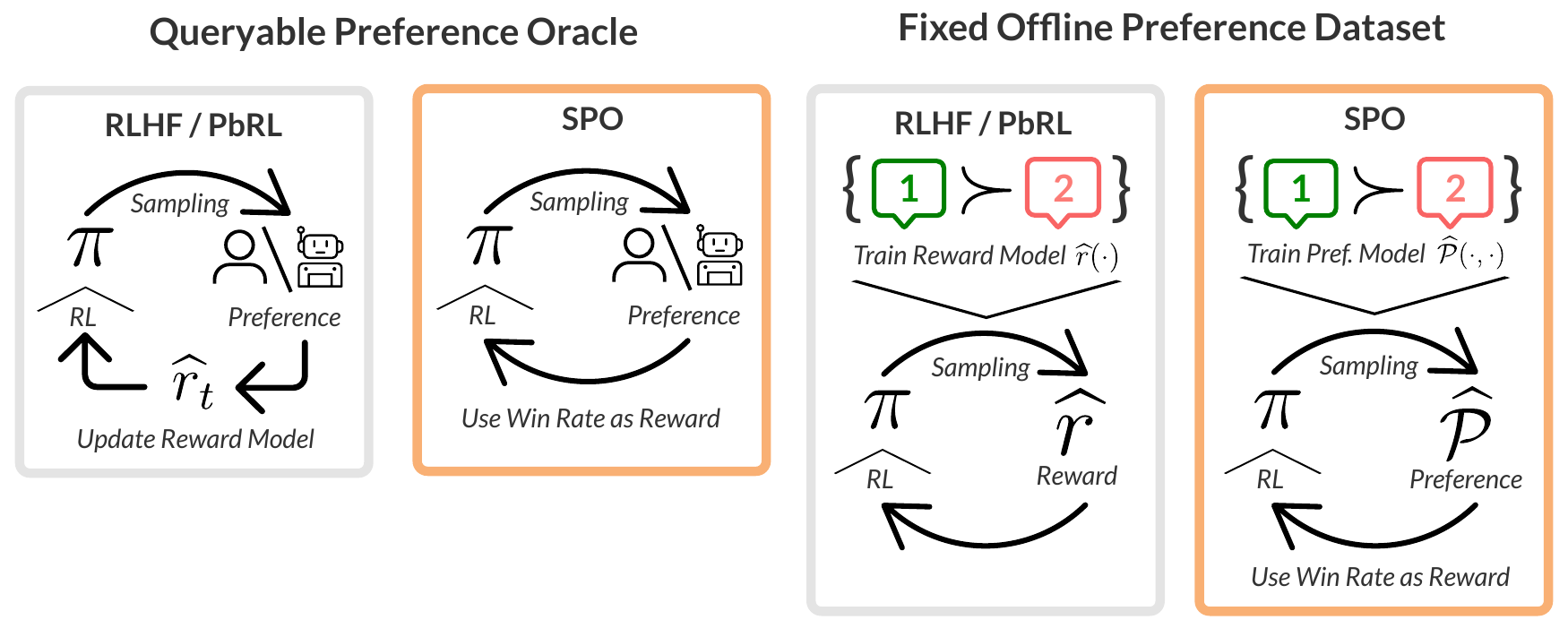}
    \caption{\footnotesize \textbf{Left}: In the setting where we are able to query the preference function online, \texttt{SPO} saves us a detour through a reward model. \textbf{Right:} In the setting where we are given a fixed dataset, we are still able to apply \texttt{SPO}. Rather than fit a reward model, we fit a \textit{preference model} $\widehat{\mathcal{P}}: \Xi \times \Xi \to [-1, 1] $ that takes in both trajectories in each preference pairs and attempts to maximize the likelihood of the preferred trajectory. Intuitively, this model is modeling \textit{relative} probabilities rather than \textit{absolute} scores. One would then use this $\widehat{P}$ as a plug-in estimate for $\mathcal{P}$ and runs \texttt{SPO} as usual. We do not explore the offline dataset paradigm in our experiments. \label{fig:wrkflw}}
\end{figure*}
\label{thm:cyclic}
\end{theorem}
\begin{sk}

Consider a three element $\Pi$. We can construct a $\mathcal{P}$, with a unique Minimax Winner of the form $[x, x, 1 - 2x]$ for $x \in (\frac{1}{3}, \frac{1}{2})$. Assume $\pi_{\text{ref}} = [\frac{1}{3},\frac{1}{3}, \frac{1}{3}]$. Then, reward-based methods can only pick a reward model that makes \textit{(a)} one option preferred to all others (resulting in some permutation of $[1, 0, 0]$), \textit{(b)} making two options preferred to the third (resulting in some permutation of $[\frac{1}{2}, \frac{1}{2}, 0]$) or \textit{(c)} makes all equally preferred (resulting in $[\frac{1}{3},\frac{1}{3}, \frac{1}{3}]$). None of these options can represent the Minimax Winner.
\end{sk}

However, by appeal to Equation \ref{eq:pirm}, standard RLHF algorithms / DPO applied \textit{iteratively} with each batch of preferences collected on-policy and a sufficiently high frequency of reward model updates may be able to compute MWs on average -- we explore this idea in our experiments

\noindent \textbf{Q2: \textit{If there exists an optimal policy / Copeland Winner for my problem, could running SPO be an inefficient way to compute it?}} 
We prove that this is not the case in a strong sense: when there exists a clearly optimal policy, our above algorithm converges to it at a \textit{fast statistical rate} -- $\widetilde{O}(\frac{1}{T})$ instead of the usual $\widetilde{O}(\frac{1}{\sqrt{T}})$ average regret. This matches the rates for UCB-style methods, previously been studied in more restricted versions of the dueling setup~\citep{JMLR:v22:18-546}. We give a simpler version of the result here, with a more general form and proof in Appendix~\ref{app:proofs:gap}.

\begin{algorithm}[t]
\begin{algorithmic}[1]
\STATE {\bfseries Input:} Learning rate $\eta$, Iterations $T$, Preference fn. $\mathcal{P}$.
\STATE {\bfseries Output:} Trained policy $\pi$.
\STATE Initialize $\pi_h(\cdot \vert \phi) = \mathsf{Unif}(\mathcal{A})$, $\forall \phi \in \Phi_h, h \in [H]$.
\FOR{$t$ in $1 \dots T$}
\STATE \algcommentlight{Preference to $\pi_t$ as reward}
\STATE Compute $r_t(\xi) = \mathbb{E}_{\xi' \sim \pi_t}\mathcal{P}(\xi, \xi')$.
\STATE Set $Q_t^h(\phi, a) = \mathbb{E}_{\xi \sim \pi_t}[r_t(\xi)|\phi_h = \phi, a_h = a]$.
\STATE Set $A_t^h(\phi, a) = Q_t^h(\phi, a) - \mathbb{E}_{a' \sim \pi_t^h(\phi)}[Q_t^h(\phi, a')]$.
\FOR{$\phi, a, h \in \Phi_h \times \mathcal{A} \times [H]$}
\STATE \algcommentlight{use no-regret algo for update}
\STATE $\pi_{t+1}^h(a \vert \phi) \propto \pi_{t}^h(a \vert \phi) \exp{(\eta A_t^h(\phi, a))}$.
\ENDFOR
\ENDFOR
\STATE {\bfseries Return } $\bar{\pi}$, trajectory-level mixture of $\pi_{1:T}$.
\end{algorithmic}
\caption{\texttt{SPO} (Theoretical Version) \label{alg:spo-theory}}
\end{algorithm}

\begin{corollary}[Informal]
Suppose that there exists $\pi^*\in\Pi$ such that $\mathcal{P}(\pi^*, \pi) > \Delta$ for all $\pi \ne \pi^*$ and $-\Delta \leq \mathcal{P}(\pi, \pi') \leq \Delta$ for all $\pi, \pi' \ne \pi^*$. Then, after $T$ rounds, the average solution $\bar{p}$ computed using Hedge as oracle $\mathcal{O}$ and $\ell_t^{\texttt{SPO}}$ is an $\widetilde{O}(\frac{|\Pi|}{\Delta T})$-approximate Minimax Winner. 
\label{corr:fast}
\end{corollary}

\subsection{\textbf{\texttt{SPO}}: Self-Play Preference Optimization.}
For single-step problems with a small and discrete policy class, it is common to maintain a distribution over policies / arms. %
However, as we transition to the sequential setting with a large and often continuous policy class, it is difficult to scale such an approach. We are therefore faced with the question of \textit{what is the right no-regret algorithm to optimize the sequence of \texttt{SPO} losses (Equation \ref{eq:pirm}) in the RL setting?}

To answer this question, we turn to the celebrated idea of \textit{local regret minimizers} \citep{zinkevich2007regret, even2009online}. Consider a problem with a finite state space. Then, at each state $s \in \mathcal{S}$, we could independently instantiate a no-regret algorithm that optimizes over $\Delta(\mathcal{A})$, feeding it a loss that depends on the cumulative reward received after exiting the state, i.e. $Q(s, a)$. Then, regardless of the state distribution our resulting policy induces, we can guarantee that we're improving at each iteration. For a specific no-regret algorithm (Hedge, \citet{freund1997decision}), this leads to the well-known \textit{soft policy iteration} (SPI) procedure \citep{ziebart2010modeling}. Because our reward function is at the trajectory level, we technically need to have a regret minimizer at each \textit{history} rather than at each state. We describe in Algorithm \ref{alg:spo-theory} an instantiation of \texttt{SPO} that uses history-dependent SPI as its policy optimizer (Lines 6-10) and now present a performance guarantee on the learned policy.
\begin{corollary} With an appropriate setting of $\eta$, running Algorithm \ref{alg:spo-theory} for T iterations guarantees that $\bar{\pi}$ is a $8 H \sqrt{\frac{\log(|\mathcal{A}|)}{T}}$-approximate Minimax Winner. \hyperref[pf:corr:spo_rl]{[Proof]}
\label{corr:spo_rl}
\end{corollary}

This follows directly from our Theorem \ref{thm:spo-ff}. 

\noindent \textbf{\texttt{SPO} in the Contextual Bandit Setting.} For certain RLHF applications (i.e. LLM preference fine-tuning), it is common to model the problem as a contextual bandit rather than a bona fide RL problem. In such settings, SPI simplifies back down to the Hedge algorithm \citep{freund1997decision}. This means that for all contexts (prompts) $x \in \mathcal{X}$ and all arms (completions) $y \in \mathcal{Y}$, \texttt{SPO} (Algorithm \ref{alg:spo-theory}) simplifies to
\begin{equation}
{\color{perfblue}
\boxed{
 \pi_{t+1}(y | x) \propto \pi_{t}(y | x) \cdot \exp{\bigg(\eta \mathbb{E}_{\substack{y' \sim \pi_t(x)}}[\mathcal{P}(y \succ y' | x)]\bigg)}. \label{eq:hedge_spo}} \nonumber}
\end{equation}
We do not explore the contextual setting in our experiments and leave a thorough study of approximations of the above update that scale to modern-day generative modeling applications to future work. See Appendix \ref{app:proofs:context} for the natural extension of Algorithm \ref{alg:spo-theory} the contextual MDP setting.

\noindent \textbf{\texttt{SPO} in Continuous Control.} 
One often uses a deep network to represent their policy rather than the tabular representation we assume in Algorithm \ref{alg:spo-theory}. It turns out that for certain policy parameterizations, the Natural Policy Gradient (NPG) algorithm of \citet{kakade2001natural} is \textit{exactly} equivalent to the soft policy iteration procedure \citep{agarwal2021theory}. Many standard deep RL algorithms like TRPO \citep{schulman2015trust} and PPO \citep{schulman2017proximal} are explicitly motivated as approximating NPG, while techniques like SAC \citep{haarnoja2018soft} can also be viewed as approximate soft policy iteration. Thus, as we move towards practice, we are free to choose from a wide set of deep RL techniques as reasonable approximations of Algorithm \ref{alg:spo-theory}.

\begin{algorithm}[t]
\begin{algorithmic}[1]
\STATE {\bfseries Input:} Iterations $T$, Preference fn. $\mathcal{P}$, Queue size $B$, Reinforcement learning algo. $\mathsf{RL}: \Pi \times \mathcal{D} \to \Pi$.
\STATE {\bfseries Output:} Trained policy $\pi$.
\STATE Initialize $\pi_1 \in \Pi$, Queue $\mathcal{Q} \gets [\xi_{1:B} \sim \pi_1 ]$.
\FOR{$t$ in $1 \dots T$}
\STATE Sample $\xi_t \sim \pi_t$.
\STATE \algcommentlight{Win-rate over queue as reward}
\STATE Compute $r_t(\xi_t) = \frac{1}{|\mathcal{Q}|}\sum_{q = 1}^B \mathcal{P}(\xi_t, \xi_q)$.
\STATE Set $r^h_t = r_t(\xi_t) / H$, $\forall h \in [H]$.
\STATE \algcommentlight{use PPO, TRPO, SAC ...}
\STATE $\pi_{t+1} \gets \mathsf{RL}(\pi_t, \mathcal{D} = \{(s^h_t, a^h_t, r^h_t)\}_{h \in [H]})$. 
\STATE $\mathcal{Q} \gets [\xi_2, \dots, \xi_B, \xi_t]$.
\ENDFOR
\STATE {\bfseries Return } best of $\pi_{1:T}$ on validation data.
\end{algorithmic}
\caption{\texttt{SPO} (Practical Version) \label{alg:spo-praxis}}
\end{algorithm}
One other concern we need to resolve is how to do credit assignment with trajectory-level feedback. In fact, if we are unable to provide rewards for each timestep in the problem, our options for policy optimization are quite limited outside of notoriously high variance REINFORCE-style policy gradients \citep{williams1992simple}. We suggest a simple fix to this problem: just split the trajectory-level reward equally amongst all state-action pairs. We prove in Appendix \ref{app:proofs} that doing so preserves policy optimality.
\begin{lemma}
Consider a trajectory-level reward function $r$ and define $\Pi^\star = \argmax_{\pi \in \Pi} \mathbb{E}_{\xi \sim \pi}[r(\xi)]$ as the corresponding set of optimal policies. For all $(s_t, a_t)$ in any $\xi$, let $\Tilde{r}(s_t, a_t) = r(\xi) / H$. Then, we have that $\Pi^\star = \argmax_{\pi \in \Pi}\mathbb{E}_{\xi \sim \pi}[\sum_h^H \Tilde{r}(s_t, a_t)]$. \hyperref[pf:lem:shap]{[Proof]}
\label{lem:shap}
\end{lemma}
In general, such a transformation can cause issues with learning good state-based critics due the fundamentally non-Markovian nature of a trajectory-level reward. However, we find that in practice, this reward transformation (Line 7 in Algorithm \ref{alg:spo-praxis}) significantly speeds up policy search. 

Lastly, in theory, \texttt{SPO} requires sampling multiple trajectories per policy update. In practice, we simply keep a queue $\mathcal{Q}$ of a small, fixed size (typically 10-100) and use the win rate against this queue for labeling trajectories sampled from the current policy (Line 6 in Algorithm \ref{alg:spo-praxis}). This makes our approach strikingly lightweight to implement on top of any policy optimization method of choice: \textit{it is just reward relabeling}. We describe our full approach in Algorithm \ref{alg:spo-praxis}. We remark that this trick of maintaining a queue only works in settings without context, and in contextual settings like LLMs, we still need to query and compare multiple actions at each context -- see Algorithm~\ref{alg:spo-praxis-contexts} for details.

\begin{figure}[t]
    \centering
    \begin{subfigure}{0.23\textwidth}
    \includegraphics[width=\textwidth]{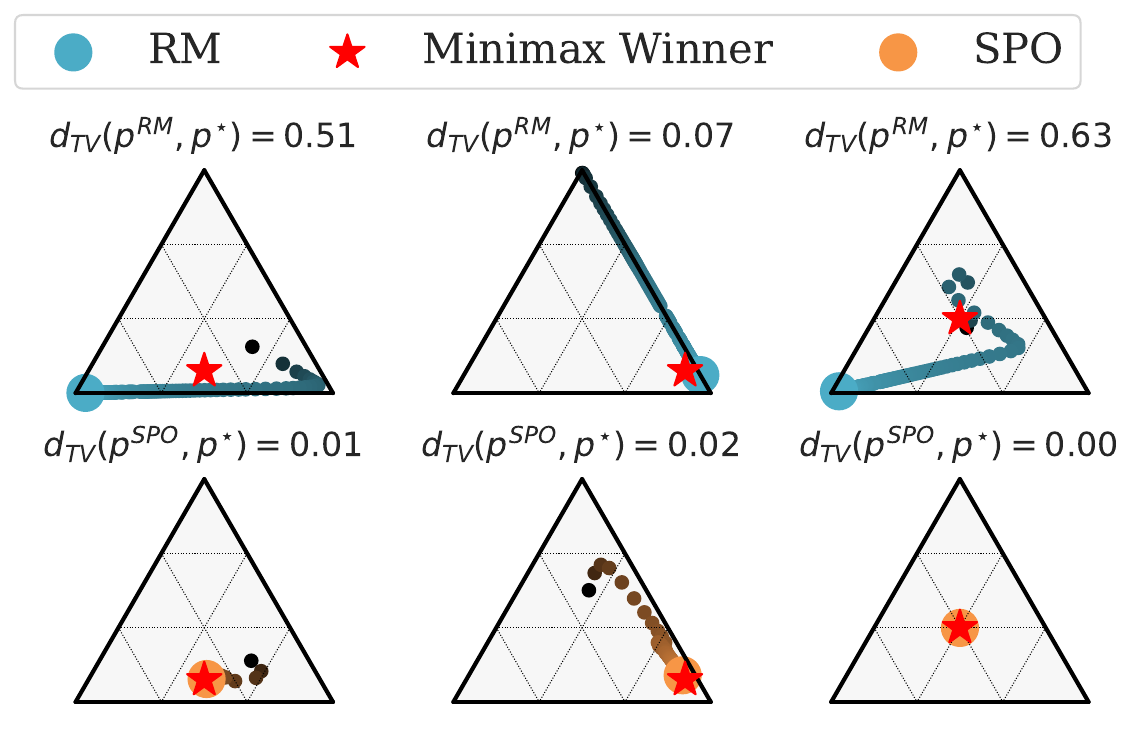}
    \caption{Intransitive Prefs. (Discrete)}
    \end{subfigure}
    \begin{subfigure}{0.22\textwidth}
    \centering
    \includegraphics[width=0.75\textwidth]{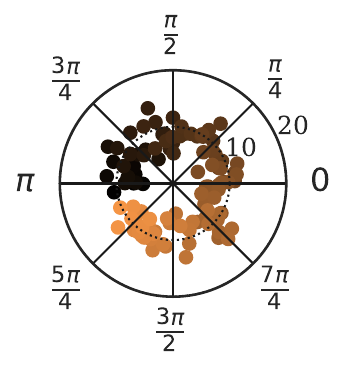}
    \caption{Intransitive Prefs. (Ant-v3)}
    \end{subfigure}
    \caption{\footnotesize\textbf{(a):} Across a variety of intransitive preferences over a discrete set of three options, \texttt{SPO} learns the MW almost exactly, while RM always converges to a corner -- see Appendix \ref{app:exps:disc} for details. \textbf{(b):} We visualize the position of our agent in a Mujoco Ant-v3 environment at the end of an episode over the course of training. Each dot represents 100k steps of training. We see that our agent traces out a circle of radius 10 (the MW) on average. More seeds and results in Fig.~\ref{fig:cyclic_extra}.}
    \label{fig:max_reward}
\end{figure}
\begin{figure*}[h]
    \centering
\begin{subfigure}{0.98\textwidth}
    \centering
    \includegraphics[width=0.245\textwidth]{figures/max_reward_pref_walker_walk.pdf}
    \includegraphics[width=0.245\textwidth]{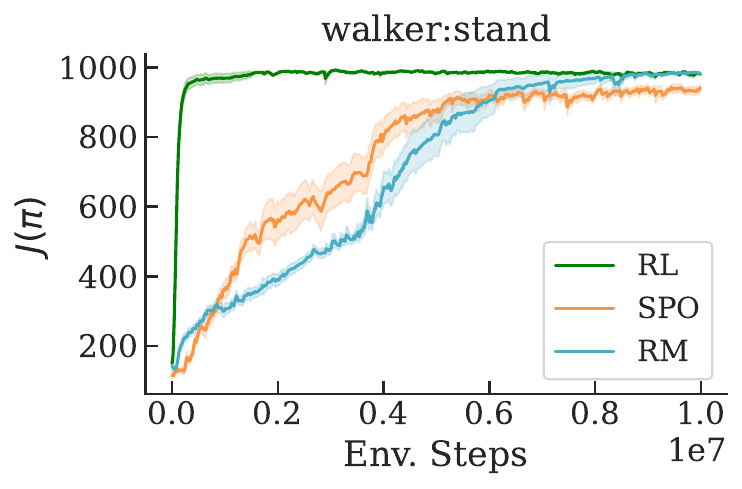}
        \includegraphics[width=0.245\textwidth]{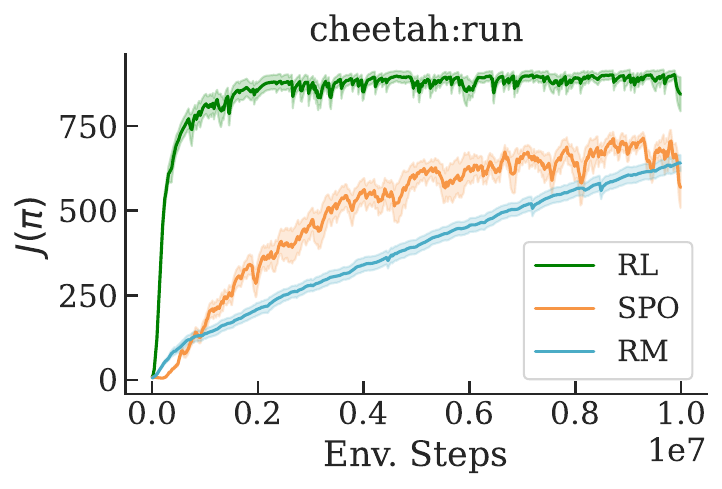}
            \includegraphics[width=0.245\textwidth]{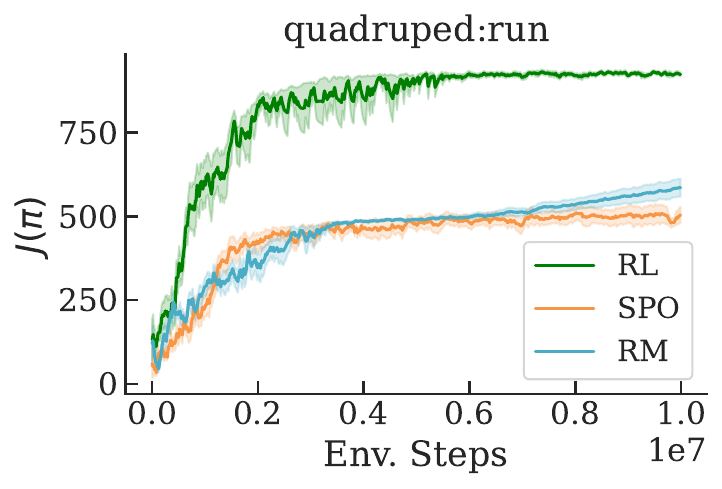}
    \caption{Maximum Reward Preferences}
    \end{subfigure}
    \begin{subfigure}{0.98\textwidth}
        \centering
    \includegraphics[width=0.245\textwidth]{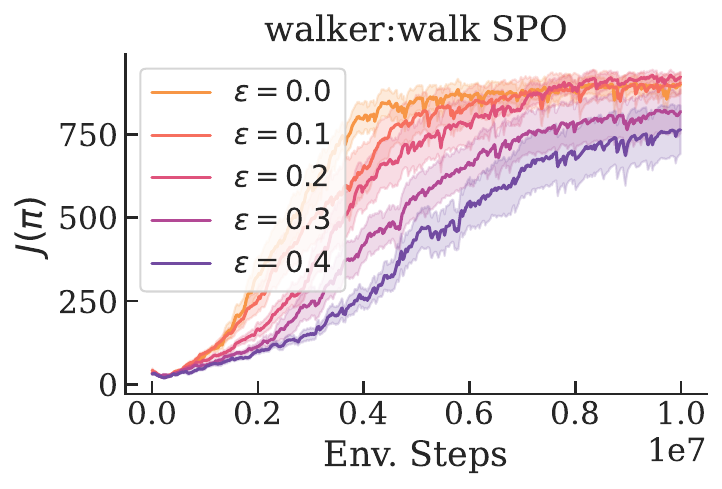}
    \includegraphics[width=0.245\textwidth]{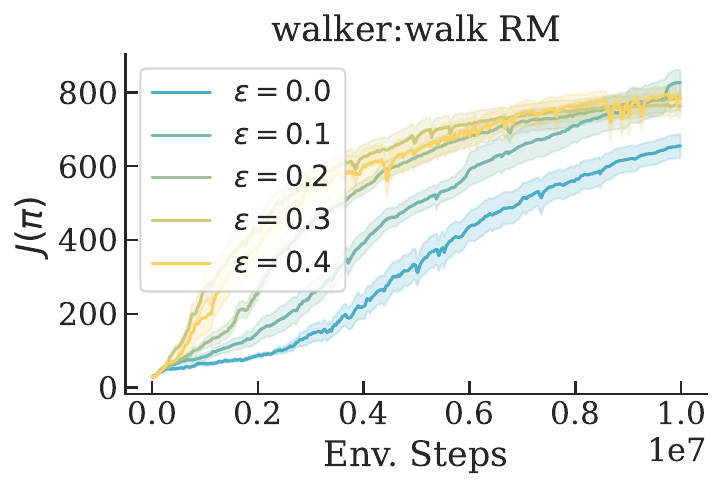}
    \includegraphics[width=0.245\textwidth]{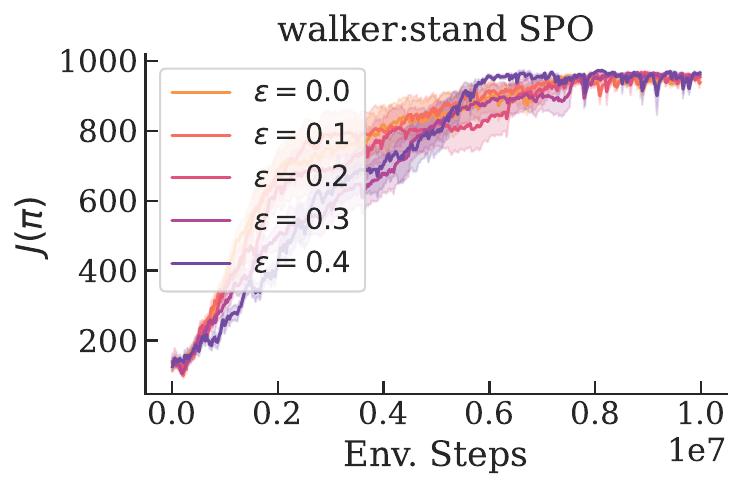}
    \includegraphics[width=0.245\textwidth]{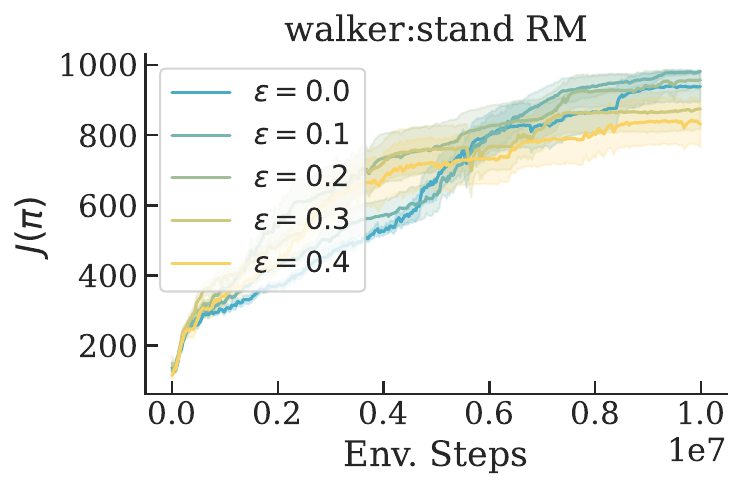}
    \caption{Stochastic Preferences}
    \end{subfigure}
    \begin{subfigure}{0.98\textwidth}
    \centering

    \includegraphics[width=0.245\textwidth]{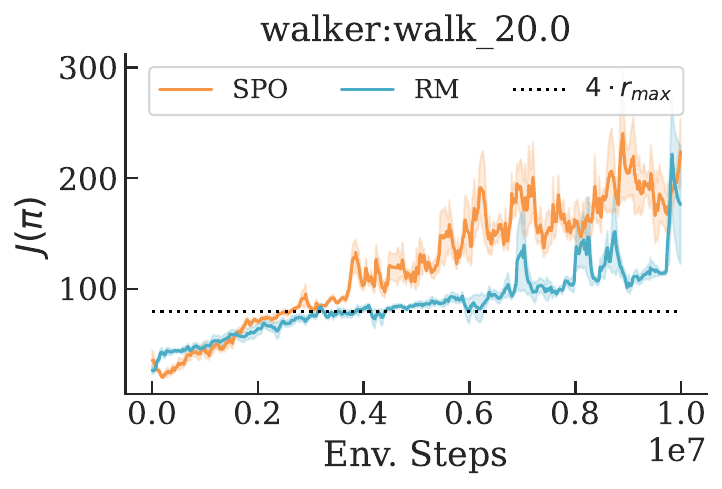}
    \includegraphics[width=0.245\textwidth]{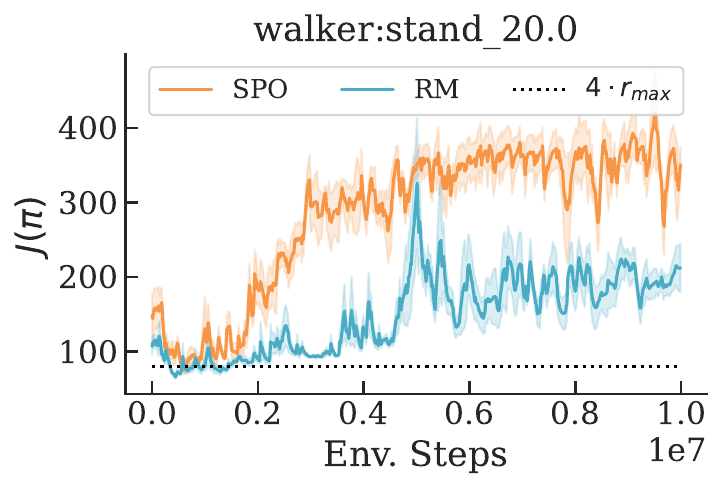}
    \includegraphics[width=0.245\textwidth]{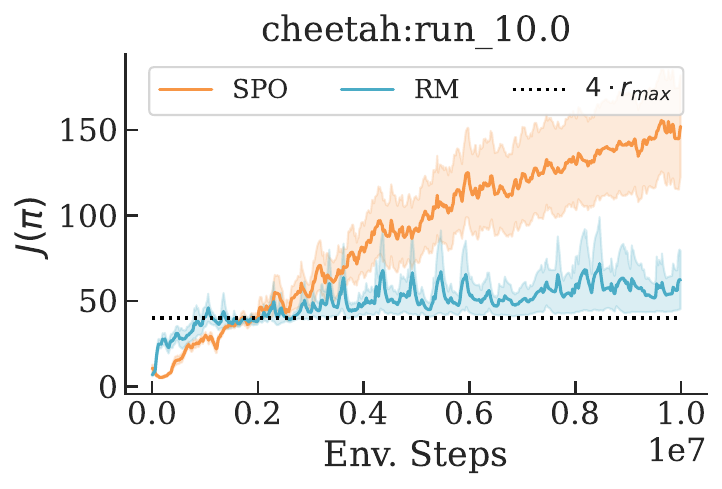}
    \includegraphics[width=0.245\textwidth]{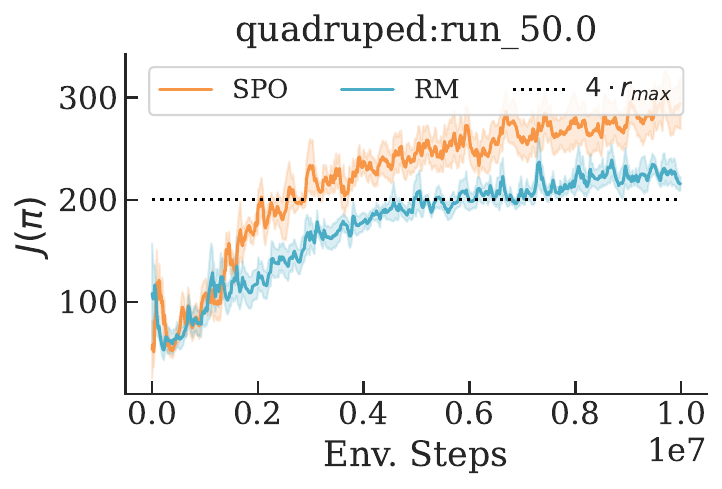}
    \caption{Non-Markovian Preferences}
    \end{subfigure}
    \caption{\footnotesize \textbf{(a):} When the preferred trajectory is that with the higher ground-truth reward, \texttt{SPO} is able to match or improve upon the RM-based approach. \textbf{(b)} We ablate the robustness of \texttt{SPO} and reward-model based approaches to preference feedback being flipped w.p. $\epsilon$. Despite the fact that \texttt{SPO} doesn't learn an extra model to average out the noisy feedback, on some environments, it is able to match the performance of RM. \textbf{(c)} We compare \texttt{SPO} and RM-based approaches on a non-Markovian task: maximize the reward subject to the constraint that the total reward accrued during the last quarter must be at most $r_{\text{max}}$. \texttt{SPO} learns policies that exploit their freedom early in the episode while RM struggles to cross the $4 \cdot r_{\text{max}}$ threshold. Standard errors are computed across 10 seeds. }
    \label{fig:noisy_feedback}
\end{figure*}

\section{Experimental Evaluation}

We evaluate \texttt{SPO} experimentally and compare it against an iterative Reward Modeling (RM) approach along several axes. We focus on the context-free, online oracle setting and leave exploration of the contextual, offline dataset setting to future work. Specifically, we ask the following questions:

\noindent \textbf{1. Can \texttt{SPO} compute MWs when faced with intransitive preferences?} We consider aggregating three populations in different proportions, each of which has transitive preferences internally. We measure how far off \texttt{SPO} is from the exact MW. We also present qualitative results on a continuous control task from Mujoco, \citep{Brockman16Gym} where computing the MW for comparison is infeasible.

\noindent \textbf{2. How sample efficient is  \texttt{SPO} on problems with unique Copeland Winners / optimal policies?} We evaluate \texttt{SPO} and RM with preferences based on ground truth rewards from the DMControl \citep{Tassa18DMControl} continuous control environments. This setting is tailor-made for RM as there exists a deterministic, Markovian reward function that explains the observed preferences.

\noindent \textbf{3. How robust is \texttt{SPO} to stochastic preferences?} We study the robustness of RM and \texttt{SPO} to corruptions of various probabilities (i.e. Bernoulli noise) in preference labels. This setup is meant to capture some of the stochasticity in human preferences that makes RLHF challenging in practice.

\noindent \textbf{4. Can \texttt{SPO} handle Non-Markovian preferences?} We consider a challenging situation where we want to elicit \textit{qualitatively} non-Markovian behavior (e.g. constraints on just a part of a trajectory) from a Markovian policy purely on the basis of  trajectory-level relative feedback.

To eliminate data staleness issues, we allow both \texttt{SPO} and RM to query the preference function online, and thus continuously update the RM during policy search. Hence, RM can be seen as a \textit{maximally iterative} reward-based method and therefore much closer to \texttt{SPO} than the traditional two-stage procedure often employed in RLHF. \footnote{In Figure \ref{fig:frozen_rm}, we observe that freezing the reward model partway through training (i.e. moving us closer to an offline RLHF recipe) leads to a consistent decline in performance for RM.} As we discussed above, reward-based methods with sufficiently high update frequencies resemble \texttt{SPO}, with the only difference potentially being the training distribution for the underlying RM, when compared with the implicit \texttt{SPO} reward~\eqref{eq:pirm}. Thus, our investigation focuses on the subtle questions: \textit{whether learning a reward model provides any benefit over directly using preference data when teacher queries are cheap.} 

We use Soft Actor Critic (SAC, \citet{haarnoja2018soft}) for continuous control and Proximal Policy Optimization (PPO, \citet{schulman2017proximal}) for discrete action tasks, both as implemented in the ACME framework \citep{hoffman2020acme}. The actor, critic sizes and activations are kept same between \texttt{SPO} and RM. Explicitly, the \textit{only} distinction between \texttt{SPO} and RM is how the reward of a trajectory is estimated and fed into the policy optimization algorithm. %

The reward model is Markovian and trained on trajectory level comparisons using trajectories drawn from the agent's replay buffer by optimizing the Bradley-Terry loss~\citep{bradley1952rank}. \footnote{As prior work often uses snippet-level feedback in this domain \citep{lee2021pebble}, we ablate its effect in Figure \ref{fig:snippet_ablation} and find that it can either help or hurt depending on the environment. We also do not include several other techniques from \citet{lee2021pebble} like reward model ensembles, active query selection, or batched rather than interleaved  updates of the reward model as our work focuses on the fundamental differences between preference-based and reward-based methods, rather than query efficiency in RLHF. We note that all of the aforementioned techniques are equally applicable to both learned reward and preference models.} During learning, the current reward model is used to label samples that are drawn from the replay buffer for performing policy search. The relabeling works better in our experiments than using the rewards assigned when these samples were put into the replay buffer. We keep the reward model architectures consistent with~\citet{lee2021pebble} and perform extensive sweeps over learning rates and update rules -- see Appendix \ref{app:exps} for more details. \textbf{We postpone additional results for all experiments to Appendix \ref{app:extra_res}.}

\noindent \textbf{Intransitive Preferences.} We begin by testing whether \texttt{SPO} is actually able to compute MWs in practice via attempting to optimize \textit{cyclic} (intransitive) preferences. 

First, we consider 3 populations, each of which has internally transitive preferences over a discrete set of 3 options. However, when aggregated, their preferences become intransitive, as is common in real-world scenarios \citep{may1954intransitivity}. Because this problem is discrete, we can compute the MW in closed form. As we show in Fig. \ref{fig:max_reward} (a) (also Figs.~\ref{fig:rps_bandit_spo}, \ref{fig:rps_bandit_rm}), \texttt{SPO} almost exactly computes the MW across a range of sub-population weightings, agreeing with Theorem \ref{thm:spo-ff}. In contrast, RM forces a total order over the actions, and the policy typically converges to a corner far from the MW.

Next, we consider the Mujoco Ant-v3 navigating atop a 2D plane and consider its radius and angle with respect to the origin at the end of the episode, $(R(\xi), \theta(\xi))$. We consider a preference where a trajectory looses to the ``pizza-slice'' of angle $\bar{\theta}=\pi/4$ in front of it and, within each slice, prefers the ``crust'' to the ``cheese''.
While we are unable to compute the exact MW here, the symmetry of the preference function over angles implies that the MW qualitatively chooses points uniformly across the slices, while keeping a minimum distance from the center. In Figures \ref{fig:max_reward} (b) and~\ref{fig:cyclic_extra}, we see that over the course of training, our agent matches this behavior on average, continuously sweeping out full circles.

\noindent \textbf{Maximum Reward Preference.} We next consider tasks where the preferences are determined according to comparisons based on the underlying ground truth reward function, i.e. $\mathcal{P}_{1}(\xi, \xi')=2\cdot\mathbbm{1}[r(\xi)>r(\xi')]-1$, where $r(\xi)=\sum_{t=1}^H r_{\text{GT}}(s^\xi_t,a^\xi_t)$, with $r_{\text{GT}}(s_t^\xi,a_t^\xi)$ denoting the ground truth reward for the $t^{\text{th}}$ state-action pair in trajectory $\xi$. Naturally, this is the most favorable setting for utilizing an RM based approach, where one might hope to leverage the generalization ability of an RM to perform careful credit assignment for sample efficient learning. However, based on Figure~\ref{fig:max_reward}(b) and~\ref{fig:max_reward_extra}, this isn't always the case. This dovetails with our Corollary \ref{corr:fast} on the fast rates enjoyed by \texttt{SPO} when preferences are explained by a reward function. %

\noindent \textbf{Noisy Preferences.} A natural followup to ground truth reward based preferences is to test the robustness of these methods to noisy preferences (representing annotator disagreements). We study this setting by flipping the maximum reward preference $\mathcal{P}_1(\cdot)$ above according to i.i.d Bernoulli noise $k$, i.e. $\mathcal{P}_2(\xi,\xi')=(1 - k) \mathcal{P}_1(\xi,\xi') + k\cdot\mathcal{P}_1(\xi',\xi)$, where $k\sim\text{Bern}(\epsilon)$. As we do not learn an extra model in our implementation of \texttt{SPO}, we would expect it to be less robust to this sort of noise due to the lack of an additional averaging effect.
In Figure \ref{fig:noisy_feedback} (a) and \ref{fig:noisy_feedback_extra}, we see that \texttt{SPO} performs comparably to RM on some environments while performing noticeably worse on some. Taken with the preceding result, this result implies that while learning a parametric model from preference feedback has limited utility when raters are always correct, doing so can provide a strong emperical benefit in some stochastic situations. Thus, an interesting question to explore in future work is whether learning an iterative \textit{preference} model helps close this gap.

\noindent \textbf{Non-Markovian Preferences.} Lastly, we consider a challenging task where we want the agent to maximize their cumulative reward as much as possible subject to the constraint that their total reward in the last quarter of a trajectory is below a threshold $r_{\text{max}}$. 
The optimal strategy for this setting is to exhibit \textit{qualitatively} non-Markovian behavior, i.e. maximize reward as much as possible during the first $\frac{3}{4}$ths of an episode before switching to more conservative behavior. The reason this is challenging is because we are optimizing over \textit{Markovian} policies, which means the agent needs to learn an unusually complex mapping. In Figs. \ref{fig:noisy_feedback} (b) and~\ref{fig:noisy_feedback_extra}, we see that while \texttt{SPO} is consistently able to cross the $4 \cdot r_{\text{max}}$ threshold that corresponds to exhibiting qualitatively non-Markovian behavior, RM often struggles to do so. %

\section{Discussion}

This paper develops SPO, a simple and effective approach to RLHF. We confirm the efficacy both in our theoretical analysis and diverse evaluation in control tasks. In theory, better understanding of last iterate convergence issues with bandit feedback, and handling uncertainty in learned preference functions are interesting research directions. A natural empirical question is applying these ideas to fine-tuning generative models with AI feedback from large models, or using preference models learned from human annotations. Relatedly, an computational limitation of preference model-based methods compared to reward model-based in the contextual setting worth exploring is the need to sample multiple times (at least twice) to compute scores for a generation.

\section*{Acknowledgments}
ZSW is supported in part by the NSF FAI Award \#1939606,
a Google Faculty Research Award, a J.P. Morgan Faculty
Award, a Facebook Research Award, an Okawa Foundation
Research Grant, and a Mozilla Research Grant. GKS would
like to thank Drew Bagnell for valuable feedback.

\section*{Impact Statement}
This paper presents work whose goal is to advance the field of Machine Learning. There are many potential societal consequences of our work, none which we feel must be specifically highlighted here.

\bibliography{example_paper}
\bibliographystyle{icml2024}

\newpage
\appendix
\onecolumn

\section{Related Work}
\label{app:rel}

\noindent \textbf{Dueling Bandits and Dueling RL.} Beginning with the seminal work of \citet{yue2012k}, various authors have viewed preference-based optimization of a multi-armed or contextual bandit as a two-player zero-sum game \citep{dudik2015contextual, saha2021adversarial, saha2022efficient, JMLR:v22:18-546}. \citet{dudik2015contextual} carry out a detailed theoretical study of the two-player approach in contextual bandits, where they use the name \emph{von Neumann winner} to refer to the concept of a Minimax Winner. We adopt the latter name due to its older roots in the social choice theory literature. More recently, various authors have investigated dueling reinforcement learning algorithms from a theoretical perspective. In contrast to \citet{pacchiano2023dueling}, we do not need to assume preferences are explained by an underlying reward function. We build upon the work of \citet{wang2023rlhf} by utilizing their reduction of RLHF to adversarial MDP solving. However, we further leverage the structure of the problem to derive single-player algorithms, while all aforementioned approaches requires adversarial training.

Both in the bandit and sequential settings, prior work has considered single-player algorithms for RLHF. However, these results require strong linearity assumptions and are only applicable in the bandit domain \citep{sui2017multi} or assume an underlying Markovian reward function \citep{novoseller2020dueling}. In contrast, we provide a reduction to no-regret online learning that allows one to plug in \textit{any} no-regret algorithm (e.g. Online Gradient Descent, \citet{zinkevich2003online}) without additional assumptions. In particular, by leveraging a recent result from \citet{wang2023rlhf}, we are able to prove that we can utilize a variation of the Natural Policy Gradient \citep{kakade2001natural, agarwal2021theory}, which practical policy gradient algorithms like PPO \citep{schulman2017proximal} and TRPO \citep{schulman2015trust} approximate, to efficiently compute Minimax Winners, bypassing the hardness result of \citet{daskalakis2020independent} by utilizing the structure of our particular game. 

Perhaps the most similar work to ours is the concurrent work of \citet{munos2023nash}. 
They derive a specific algorithm, focus on quantal response equilibria to be able to prove last-iterate convergence, and treat the problem as a normal-form game. Instead, we focus on a general algorithmic framework, provide convergence guarantees to the Nash equilibrium, and account for the sequential nature of the game.
Empirically, they focus on a particular task of learning document summarization from human feedback, while we study continuous control tasks where we experiment with a range of realistic preference functions. Recent work by \citet{chen2024selfplay} formulates inverse RL for LLM fine-tuning as a kind of self-play -- we focus on optimizing from preferences rather than from demonstrations.

Our reward model baseline and our continuous control setup are heavily influenced by the works of \citet{christiano2017deep, lee2021pebble}. One critical distinction from this prior work is rather than assuming that the rater is able to provide snippet-level feedback, we only assume they can perform trajectory-level comparisons, a much less dense form of supervision. 

\noindent \textbf{RLHF without Reward Models.} Recently, several authors have proposed eliminating reward models from RLHF by leveraging the well-known bijection between the optimal policies of minimum-relative-entropy RL problems and their advantage functions \citep{ziebart2010modeling} to directly optimize the policy by substituting it into the classification loss usually used to train the reward model \citep{zhao2023slic, rafailov2023direct, hejna2023contrastive, azar2023general}. While these approaches are appealing for their conceptual simplicity and ease of implementation, they suffer from the same issues with intransitive and noisy preferences as they are derived on the basis of an \textit{implicit} reward model. Additionally, as we discuss further in Appendix \ref{app:compounding_errs}, they can suffer from \textit{compounding errors} \citep{ross2011reduction} due to their offline nature. In a sense, this family of approaches can be thought of as the preference-based analog to \textit{behavioral cloning} techniques \citep{pomerleau1988alvinn} in imitation learning. \footnote{While preferences are collected on the initial policy's distribution (unlike in BC), this data becomes off-policy after policy optimization.} In contrast, the technique we propose is the preference-based analog of inverse reinforcement learning approaches \citep{ziebart2010modeling}. We summarize this taxonomy in Table \ref{onetabletorulethemall}.

\noindent \textbf{Self-Play In Language Modeling.} Various works have explored self-play for preference fine-tuning LLMs. Conceptually, this can be thought of as an instantiation of the general \texttt{SPO} reduction on a contextual bandit problem, using a variety of base learners to approximate the idealized update we propose. Concurrent to our work, \citet{munos2023nash} propose using policy gradients to optimize a family of objectives, including the \texttt{SPO} losses. Subsequent to our work, \citet{calandriello2024human} propose using Online IPO, \citet{rosset2024direct} propose using Online DPO, and \citet{gao2024rebel} propose using Online REBEL to optimize the sequence of \texttt{SPO} losses. We do not explore the contextual nor preference fine-tuning settings in our experiments and refer interested readers to the above papers for details on scalable approximations of \texttt{SPO}.

\section{Frequency of Querying Online Preference Oracle.} 
Our presentation assumed access to a preference function $\mathcal{P}$ which can be queried at each round with new trajectories. This is natural when preference labels are generated by a model learned from a previously collected preference dataset, as in recently studied RLAIF settings~\cite{bai2022training, zhao2023slic}. However, in other settings, it is desirable to directly obtain the preference labels from humans or other expert models (RLHF, \citet{tucker2020preference}), where online querying is not possible. \texttt{SPO} is compatible with \emph{batched queries} the queries in these settings. Specifically, we can freeze the policy for some batch size $B$, accumulate a dataset of $B$ trajectory pairs to compare, and then query the labels for all of them. This corresponds to mini-batching in the no-regret algorithm being used by \texttt{SPO}, which preserves no-regret for $B < O(\sqrt{T})$.

\clearpage
\section{Proofs}
\localtableofcontents

\label{app:proofs}

\subsection{Proof of Lemma \ref{lem:sym}}
\label{pf:lem:sym}
\begin{pf}
We follow a similar proof strategy to \citet{fey2012symmetric}. For convenience, we interpret the preference $\mathcal{P}$ as a matrix of size $|\Pi| \times |\Pi|$ and will write things in matrix notation, i.e. $\mathbb{E}_{\pi_1 \sim p, \pi_2 \sim q}[\mathcal{P}(\pi_1, \pi_2)] = p^T \mathcal{P} q$. We consider some $ (\hat{p}, \hat{q}) \in \mathsf{MW}(\mathcal{P})$ and will show that also $ (\hat{q}, \hat{p}) \in \mathsf{MW}(\mathcal{P})$.
Since $ (\hat{p}, \hat{q}) \in \mathsf{MW}(\mathcal{P})$ and by the definition of a Nash equilibrium,
\begin{align*}
    &\quad \max_{q \in \Delta(\Pi)} \hat{p}^T \mathcal{P} q \leq \min_{p \in \Delta(\Pi)} p^T \mathcal{P} \hat{q}  \\
    &\Rightarrow \max_{q \in \Delta(\Pi)} q^T \mathcal{P}^T \hat{p} \leq \min_{p \in \Delta(\Pi)} \hat{q}^T \mathcal{P}^T p  \\
    &\Rightarrow \max_{q \in \Delta(\Pi)} -q^T \mathcal{P} \hat{p} \leq \min_{p \in \Delta(\Pi)} -\hat{q}^T \mathcal{P} p \tag{By the anti-symmetry of $\mathcal{P}$} \\
    &\Rightarrow -\min_{q \in \Delta(\Pi)} q^T \mathcal{P} \hat{p} \leq -\max_{p \in \Delta(\Pi)} \hat{q}^T \mathcal{P} p  \\
    &\Rightarrow \min_{q \in \Delta(\Pi)} q^T \mathcal{P} \hat{p} \geq \max_{p \in \Delta(\Pi)} \hat{q}^T \mathcal{P} p  \\
    &\Rightarrow   \max_{q \in \Delta(\Pi)} \hat{q}^T \mathcal{P} q \leq \min_{p \in \Delta(\Pi)} p^T \mathcal{P} \hat{p} 
\end{align*}
Thus, $(\hat{q}, \hat{p})$ also forms a Nash equilibrium. Then, because of the interchangeability of Nash equilibrium strategies for two-player zero-sum games \citep{nash1951non}, we have that $(\hat{p}, \hat{p})$ and $(\hat{q}, \hat{q})$ are symmetric MWs.
\end{pf}

\subsection{Proof of Theorem \ref{thm:spo-ff}}
\label{pf:thm:spo-ff}
\begin{pf} We follow the strategy outlined in our preceding proof sketch. 

Consider two players, $p, q \in \Delta(\Pi)$. An $\epsilon$-approximate Nash equilibrium is a pair of strategies $(p, q)$ such that
\begin{equation}
    \max_{p^\star \in \Delta(\Pi)} \mathbb{E}_{\pi \sim p^\star, \pi' \sim q}[\mathcal{P}(\pi, \pi'))] - \min_{q^\star \in \Delta(\Pi)} \mathbb{E}_{\pi \sim p, \pi' \sim q^\star}[\mathcal{P}(\pi, \pi'))] \leq \epsilon.
\end{equation}
We define the following per-round losses for both players:
\begin{equation}
    \ell_t^1(p) = \mathbb{E}_{\pi \sim p, \pi' \sim q_t}[-\mathcal{P}(\pi, \pi')], \quad \ell_t^2(q) = \mathbb{E}_{\pi \sim p_t, \pi' \sim q}[\mathcal{P}(\pi, \pi')].
\end{equation}
We can then define the (static) regret suffered by both players as
\begin{equation}
    \mathsf{Reg}_p(T) = \sum_t^T \ell_t^1(p_t) - \min_{p^\star \in \Delta(\Pi)} \sum_t^T \ell_t^1(p^\star), \quad \mathsf{Reg}_q(T) = \sum_t^T \ell_t^2(q_t) - \min_{q^\star \in \Delta(\Pi)} \sum_t^T \ell_t^2(q^\star).
\end{equation}

By construction, we set $p_0 = q_0$. This implies that
\begin{align}
    \ell_0^1(\sigma) &=  \mathbb{E}_{\pi \sim \sigma, \pi' \sim q_0}[-\mathcal{P}(\pi, \pi')] \\
    &= \mathbb{E}_{\pi \sim \sigma, \pi' \sim p_0}[-\mathcal{P}(\pi, \pi')] \\
    &= \mathbb{E}_{\pi \sim p_0, \pi' \sim \sigma}[\mathcal{P}(\pi, \pi')] \tag{By the anti-symmetry of $\mathcal{P}$} \\
    &= \ell_0^2(\sigma)
\end{align}
If, at some time $\tau$, $p_{\tau}$ and $q_{\tau}$ have the same strategy and perform updates based on the same loss function, $p_{\tau+1} = q_{\tau+1}$.\footnote{This is trivially true if $\mathcal{O}$ is deterministic. If $\mathcal{O}$ is a randomized no-regret algorithm like Follow the Perturbed Leader \citep{kalai2005efficient}, we can instead simply share the randomness between the two players.} Then, by the anti-symmetry of $\mathcal{P}$, we have that $\ell_{\tau+1}^1 = \ell_{\tau+1}^2$. Therefore, by induction, we have that $p_t = q_t$, $\forall t \in [T]$.

We complete the proof by following the argument in \citet{freund1997decision}:
\begin{align}
\frac{\mathsf{Reg}_p(T) + \mathsf{Reg}_q(T)}{T} &= \frac{1}{T} \left(\sum_t^T \ell_t^1(p_t) - \min_{p^\star \in \Delta(\Pi)} \sum_t^T \ell_t^1(p^\star) + \sum_t^T \ell_t^2(q_t) - \min_{q^\star \in \Delta(\Pi)} \sum_t^T \ell_t^2(q^\star) \right) \\
&= \frac{1}{T} \left( - \min_{p^\star \in \Delta(\Pi)} \sum_t^T \ell_t^1(p^\star) - \min_{q^\star \in \Delta(\Pi)} \sum_t^T \ell_t^2(q^\star) \right) \tag{$\ell_t^1(p_t) + \ell_t^2(q_t) = 0$} \\
&= \frac{1}{T} \left( - \min_{p^\star \in \Delta(\Pi)} \sum_t^T \mathbb{E}_{\pi \sim p^\star, \pi' \sim q_t}[-\mathcal{P}(\pi, \pi'))] - \min_{q^\star \in \Delta(\Pi)} \sum_t^T \mathbb{E}_{\pi \sim p_t, \pi' \sim q^\star}[\mathcal{P}(\pi, \pi'))] \right) \\
&= \left(\max_{p^\star \in \Delta(\Pi)} \mathbb{E}_{\pi \sim p^\star, \pi' \sim \overline{q}}[\mathcal{P}(\pi, \pi'))] - \min_{q^\star \in \Delta(\Pi)} \mathbb{E}_{\pi \sim \overline{p}, \pi' \sim q^\star}[\mathcal{P}(\pi, \pi'))] \right).
\end{align}

Thus, $(\overline{p}, \overline{q}) = (\overline{p}, \overline{p})$ is a symmetric $\frac{\mathsf{Reg}_p(T) + \mathsf{Reg}_q(T)}{T} = \frac{2\mathsf{Reg}_p(T)}{T}$-approximate Nash equilibrium / Minimax Winner.
\end{pf}

\subsection{Proof of Theorem \ref{thm:cyclic}}
\label{pf:thm:cyclic}
\begin{pf}
Our preceding proof sketch ignored the effect of regularization for simplicity. We now provide a specific example under which standard algorithms do not compute Minimax Winners, even with regularization to a prior.

 We set $\pi_{\text{ref}}$ to be uniform to remove any trivial failures due to a lack of data support and consider the following preference matrix:
\begin{figure}[!h]
\centering
\begin{tikzpicture}
\matrix [matrix of math nodes,
         nodes={rectangle, 
                minimum size=1.75em, text depth=0.25ex,
                inner sep=0pt, outer sep=0pt,
                anchor=center},
         column sep=-0.5\pgflinewidth,
         row sep=-0.5\pgflinewidth,
         inner sep=0pt,
         ] (m)
{
  & a & b & c &  \\
a & 0 & {\color{expert}\frac{2}{5}} & {\color{error}-1} \\
b & {\color{error}-\frac{2}{5}} & 0 & {\color{expert}+1} \\
c & {\color{expert}+1} & {\color{error}-1} & 0 \\
};
\begin{scope}[on background layer]
    \filldraw[lightgray!40, rounded corners] (m-1-2.north west) -- 
        (m-1-2.south west) -- (m-1-4.south east)-- (m-1-4.north east)-- 
        cycle;
    \filldraw[lightgray!40, rounded corners] (m-2-1.north west) -- 
        (m-2-1.north east) -- (m-4-1.south east)-- (m-4-1.south west)-- 
        cycle;
\end{scope} 
\end{tikzpicture}
\caption{A preference function over $\mathcal{Y} = (a, b, c)$ with unique Minimax Winner $(\frac{5}{12}, \frac{5}{12}, \frac{1}{6})$ and unique Copeland winner $b$.}
\end{figure}

We begin by considering standard RLHF algorithms. First, we would fit a reward model via the standard Bradley-Terry loss. This would peak at the Copeland Winner, which is $b$ for the above matrix. Without loss of generality, we assume reward model outputs are in the range $[0, 1]$. Thus,

\begin{equation}
    r^\star = [0, 1, 0].
\end{equation}

From \citet{ziebart2010modeling}, we have that
\begin{equation}
    \pi^\star(y) \propto \pi_{\text{ref}}(y) \exp \left( \frac{1}{\beta} r^\star(y) \right) = \left[\frac{1}{3}, \frac{1}{3}, \frac{1}{3} \right] \odot \left[1, \exp(\frac{1}{\beta}), 1 \right] \propto  \left[1, \exp(\frac{1}{\beta}), 1 \right].
\end{equation}
Thus, for any finite non-negative $\beta$, $\pi^\star$ plays $a$ and $c$ equally often, which means it cannot play the Minimax Winner.

Next, we consider DPO. From Eq. 6 of \citet{rafailov2023direct}, we have that the optimal DPO policy has the form 
\begin{equation}
    p^\star(y_1 \succ y_2) = \frac{1}{1 + \exp{\left(\beta \log \frac{\pi^\star(y_2)}{\pi_{\text{ref}}(y_2)} - \beta \log \frac{\pi^\star(y_1)}{\pi_{\text{ref}}(y_1)} \right)}}.
\end{equation}
As $\pi_{\text{ref}}$ is uniform, we have that $\pi_{\text{ref}}(y_2) = \pi_{\text{ref}}(y_1)$ and thus we can simplify our above expression
\begin{align}
    p^\star(y_1 \succ y_2) &= \frac{1}{1 + \exp{\left(\beta \log \pi^\star(y_2) - \beta \log \pi^\star(y_1) \right)}} \\
    &= \frac{1}{1 + \left(\frac{\pi^\star(y_2)}{\pi^\star(y_1)}\right)^{\beta}} \\
    &= \frac{\pi^\star(y_1)^{\beta}}{\pi^\star(y_1)^{\beta} + \pi^\star(y_2)^{\beta}}.
\end{align}
As written, the DPO loss assumes unweighted preferences (i.e. it assumes that all positive samples are equally preferable to their corresponding negative samples). We therefore perform the natural transformation from log likelihood to cross entropy: 
\begin{align}
    \ell_{\text{DPO}}(\pi) &= - \sum_{y_1, y_2 \in \mathcal{Y} \times \mathcal{Y}} \left[\mathcal{P}(y_1 \succ y_2) \log \left( p^\star(y_1 \succ y_2) \right) \right] \\
    &= - \sum_{y_1, y_2 \in \mathcal{Y} \times \mathcal{Y}} \left[\frac{\mathcal{P}(y_1, y_2) + 1}{2} \log \left( \frac{\pi(y_1)^{\beta}}{\pi(y_1)^{\beta} + \pi(y_2)^{\beta}} \right) \right] \\
\end{align}
Via Gibbs' inequality, we know that cross-entropy is minimized when the two distributions are equal. We can then plug in the values from our above preference matrix to arrive at the following set of constraints:
\begin{equation}
    \frac{7}{10} = \frac{\pi^\star(a)^{\beta}}{\pi^\star(a)^{\beta} + \pi^\star(b)^{\beta}}, \quad 1 = \frac{\pi^\star(b)^{\beta}}{\pi^\star(b)^{\beta} + \pi^\star(c)^{\beta}}, \quad 1 = \frac{\pi^\star(c)^{\beta}}{\pi^\star(a)^{\beta} + \pi^\star(c)^{\beta}}.
\end{equation}
Clearly, it is impossible to simultaneously satisfy all of these constraints. Thus, DPO is unable to learn the minimizer of the preference-level cross-entropy loss function because of its assumption of an implicit reward model. Unfortunately, this makes analyzing the solution DPO would actually pick rather difficult from a theoretical perspective. In response, we show that regardless of the setting of $\beta$, there exists another strategy $(\pi_{\text{ref}})$ with a lower loss than the Minimax Winner.

We can write out the above loss more explicitly as
\begin{align}
    \ell_{\text{DPO}}(\pi) &= - \left(\frac{7}{10} \log \left( \frac{\pi(a)^{\beta}}{\pi(a)^{\beta} + \pi(b)^{\beta}} \right) + \frac{3}{10} \log \left( \frac{\pi(b)^{\beta}}{\pi(a)^{\beta} + \pi(b)^{\beta}} \right)  + \log \left( \frac{\pi(b)^{\beta}}{\pi(b)^{\beta} + \pi(c)^{\beta}} \right) + \log \left( \frac{\pi(c)^{\beta}}{\pi(a)^{\beta} + \pi(c)^{\beta}} \right) \right) \nonumber
\end{align}

First, observe that plugging in $\pi_{\text{ref}}$ gives us a loss value $\log(2)$ for any value of $\beta$. Also, observe that $\lim_{\beta \to 0} \ell_{\text{DPO}}(\pi_{\text{MW}}) = \log(2)$. Taking the derivative with respect to $\beta$, we get
\begin{align}
    \nabla_{\beta} \ell_{\text{DPO}}(\pi) &= \frac{-7}{10} \log \left( \frac{\pi(a)}{\pi(b)}\right) \frac{\pi(b)^{\beta}}{\pi(a)^{\beta} + \pi(b)^{\beta}} 
    \\ &+ \frac{-3}{10} \log \left( \frac{\pi(b)}{\pi(a)}\right) \frac{\pi(a)^{\beta}}{\pi(b)^{\beta} + \pi(a)^{\beta}} 
    \\ &+ -1 \log \left( \frac{\pi(b)}{\pi(c)}\right) \frac{\pi(c)^{\beta}}{\pi(c)^{\beta} + \pi(b)^{\beta}} 
    \\ &+ -1 \log \left( \frac{\pi(c)}{\pi(a)}\right) \frac{\pi(a)^{\beta}}{\pi(a)^{\beta} + \pi(c)^{\beta}}.   \nonumber
\end{align}
Now, plugging in the Minimax Winner, we get that
\begin{align}
    \nabla_{\beta} \ell_{\text{DPO}}(\pi_{\text{MW}}) &=  - \log \left( \frac{5}{2}\right) \frac{\pi_{\text{MW}}(c)^{\beta}}{\pi_{\text{MW}}(c)^{\beta} + \pi_{\text{MW}}(b)^{\beta}}  - \log \left( \frac{2}{5}\right) \frac{\pi_{\text{MW}}(a)^{\beta}}{\pi_{\text{MW}}(a)^{\beta} + \pi_{\text{MW}}(c)^{\beta}} > 0.   \nonumber
\end{align}
Thus, $\forall \beta \in (0, \infty)$, $\ell_{\text{DPO}}(\pi_{\text{MW}}) > \ell_{\text{DPO}}(\pi_{\text{ref}})$. Thus, DPO will never pick the Minimax Winner.

\end{pf}
\subsection{Proof of Corollary \ref{corr:spo_rl}}
\label{pf:corr:spo_rl}
\begin{pf}
    We consider optimization over the full history-dependent policy class. That is, for each $\pi \in \Pi$, $\pi = (\pi_1, \dots, \pi_H)$, where $\pi_h \in \Pi_h =  \{\Phi_h \to \Delta(\mathcal{A})\}$. As $\Delta(\mathcal{A})$ is convex and compact, so is $\Pi_h$ (as the set of functions to a convex set is convex), which means $\Pi$ is as well (as Cartesian products preserve convexity). We therefore satisfy the conditions for the application of Theorem \ref{thm:spo-ff}. We proceed by bounding the regret of our policy selection strategy. Define $J(\pi, r_t^{\texttt{SPO}}) = \mathbb{E}_{\xi \sim \pi}[r_t^{\texttt{SPO}}(\xi)]$ as the performance of the policy $\pi$ under the induced trajectory-level reward, $r_t^{\texttt{SPO}}$ (Eq. \ref{eq:pirm}). Then,
    \begin{align}
        \mathsf{Reg}(T) &= \max_{\pi \in \Pi} \sum_{t=1}^T \ell_t^{\texttt{SPO}}(\pi_t) - \ell_t^{\texttt{SPO}}(\pi) \\
        &= \max_{\pi \in \Pi} \sum_{t=1}^T J(\pi, r_t^{\texttt{SPO}}) - J(\pi_t, r_t^{\texttt{SPO}}) \\
        &= \max_{\pi \in \Pi} H \mathbb{E}_{\substack{h \sim \mathsf{Unif}([H]) \\ \phi \sim \rho^h_{\pi}}} \left[ \sum_{t=1}^T Q^h_t(\phi, \pi) - Q^h_t(\phi, \pi_t) \right] \tag{By the finite-horizon PDL, \citet{bagnell2003policy}}
    \end{align}
    In the last step of the proof, we apply \citet{bagnell2003policy}'s finite horizon variant of \citet{kakade2002approximately}'s Performance Difference Lemma (PDL) to the history-based MDP.
     The no-regret policy update in Line~9 of the Algorithm~\ref{alg:spo-theory} is equivalent to running Hedge \citep{freund1997decision} with $\ell_t^{\phi, h}(a) = \frac{1}{2}(1 - Q^h_t(\phi, a)) \in [0, 1]$ as loss function for all $h \in [H]$ and $\phi \in \Phi_h$ ($A_h^t$ and $Q_h^T$ are interchangeable as they differ by a per-state constant and we're taking a softmax). Thus, by the regret bound of Hedge, we have that for any comparator policy $\pi \in \Pi$, $h \in [H]$, and $\phi \in \Phi_h$,
    \begin{equation}
        \sum_{t=1}^T Q^h_t(\phi, \pi) - Q^h_t(\phi, \pi_t) \leq 4 \sqrt{\log(|\mathcal{A}|) T}.
    \end{equation}
    We can then apply this bound history-wise to our preceding expression to arrive at overall regret bound
    \begin{equation}
        \mathsf{Reg}(T) \leq 4 H \sqrt{\log(|\mathcal{A}|) T}.
    \end{equation}
    Thus, by our Theorem \ref{thm:spo-ff}, we have that $\bar{\pi}$ (the trajectory-level mixture of $\pi_{1:T}$) is a $8 H \sqrt{\log(|\mathcal{A}|) / T}$-approximate MW.
 
\end{pf}

\subsection{Proof of Lemma \ref{lem:shap}}
\label{pf:lem:shap}
\begin{pf}
Observe that for any $\xi \in (\mathcal{S} \times \mathcal{A}) ^ H$, $r(\xi) = \sum_{h=1}^H \Tilde{r}(s_t, a_t)$. Thus, $\forall \pi \in \Pi$, 
\begin{align}
    \mathbb{E}_{\xi \sim \pi}\left[r(\xi)\right] = \mathbb{E}_{\xi \sim \pi}\left[\sum_{h=1}^H \Tilde{r}(s_t, a_t)\right]
    \Rightarrow \argmax_{\pi \in \Pi} \mathbb{E}_{\xi \sim \pi}\left[r(\xi)\right] = \argmax_{\pi \in \Pi}\mathbb{E}_{\xi \sim \pi}\left[\sum_{h=1}^H \Tilde{r}(s_t, a_t)\right].
\end{align}
\end{pf}

\subsection{Proof of Corollary \ref{corr:fast}}
\label{app:proofs:gap}
We begin by stating the core assumption we will use in this section.
\begin{assumption}[Gap Condition]
    There is a subset $\Pi^\star \subseteq \Pi$ such that:
    \begin{enumerate}
        \item $\forall \pi^\star \in \Pi^\star$, $\pi \in \Pi/\Pi^\star$, $\mathcal{P}(\pi^\star, \pi) \geq \Delta$.
        \item $\forall \pi^\star_1, \pi^\star_2 \in \Pi^\star$, $- \Delta/2 \leq \mathcal{P}(\pi^\star_1, \pi^\star_2) \leq \Delta/2$.
        \item $\forall \pi_1, \pi_2 \in \Pi/\Pi^\star$, $-\Delta \leq \mathcal{P}(\pi_1, \pi_2) \leq \Delta$.
    \end{enumerate}
    \label{ass:gap-set}
\end{assumption}
Under this assumption, we can prove that rather than the $\widetilde{O}(\frac{1}{\sqrt{T}})$ rate we usually get for Hedge, we instead get a $\widetilde{O}(\frac{1}{T})$ rate.

\begin{corollary}
Under Assumption \ref{ass:gap-set}, after $T$ calls to Hedge, $\bar{p}$ is an $\frac{1 + 2|\Pi| \ln T}{\Delta T}$-approximate Minimax Winner.
\label{corr:fast2}
\end{corollary}

\begin{pf}
We start by observing that under Assumption~\ref{ass:gap-set}, the loss function $\ell_t^1(p) = \mathbb{E}_{\pi \sim p, \pi' \sim q_t}[-\mathcal{P}(\pi, \pi'))]$ observed by the minimizing player satisfies the following two properties:
\begin{enumerate}
    \item For an policy $\pi \in \Pi^\star$, we have $\ell_t(\pi) \leq \Delta$. 
    \item For any policy $\pi \in \Pi^\star$ and $\pi' \notin \Pi^\star$, $\ell_t(\pi') \geq \ell_t(\pi) + q_t(\Pi^\star)\Delta/2$, where $q_t(\Pi^\star)$ is the probability of the max player choosing policies from the set $\Pi^\star$.
\end{enumerate}

The second property follows from the fact that the loss of any policy $\pi' \notin \Pi^\star$ is always larger than the loss of any action $\pi^\star \in \Pi^\star$ by Assumption~\ref{ass:gap-set}, and the two differ by at least $\Delta/2$ whenever the comparator policy comes from the set $\Pi^\star$, since:
\begin{align*}
    \ell_t(\pi') &= \sum_{\pi'' \in \Pi} q_t(\pi'') \left[-\mathcal{P}(\pi', \pi'') \right] = \sum_{\pi'' \in \Pi^\star} q_t(\pi'') \left[-\mathcal{P}(\pi', \pi'') \right] + \sum_{\pi'' \in \Pi/\Pi^\star} q_t(\pi'') \left[-\mathcal{P}(\pi', \pi'') \right]\\
    &\geq \sum_{\pi'' \in \Pi^\star} q_t(\pi'') \Delta + \sum_{\pi'' \in \Pi/\Pi^\star} q_t(\pi'') (-\Delta) \tag{\ref{ass:gap-set}, (1) and \ref{ass:gap-set}, (3), resp.}\\
    &\geq \sum_{\pi'' \in \Pi^\star} q_t(\pi'') \left[\frac{\Delta}{2} - \mathcal{P}(\pi^\star, \pi'') \right] + \sum_{\pi'' \in \Pi/\Pi^\star} q_t(\pi'') \left[- \mathcal{P}(\pi^\star, \pi'') \right] \tag{\ref{ass:gap-set}, (2) and \ref{ass:gap-set}, (1), resp.}\\
    &= \ell_t(\pi^\star) + q_t(\Pi^\star)\Delta/2.
\end{align*}

Now let us consider the class of Follow The Regularized Leader (FTRL, \citet{mcmahan2011follow}) algorithms, which induce probability distributions as 
\begin{equation}
    p_t = \argmin_{p \in \Delta(\Pi)} \eta L_{t-1}^Tp + R(p),
    \label{eq:ftrl}
\end{equation}
for some convex regularizer $R$, and where we define $L_t = \sum_{s=1}^t \ell_s$. As $p_t$ minimizes the above expression, we know it satisfies the first-order optimality conditions, which imply that for any other distribution $p\in \Delta(\Pi)$,
\begin{equation}
    \langle \nabla R(p_t) + \eta L_{t-1}, p - p_t \rangle \geq 0.
\end{equation}
To proceed further, we consider coordinate-wise separable regularizers, that is $R(p) = \sum_{i=1}^{|\Pi|} R_i(p_i)$, and use the distribution $p = p_t + \alpha e_{\pi_a} - \alpha e_{\pi_b}$, for actions $\pi_a \in \Pi^\star$ and $\pi_b \notin \Pi^\star$, with $\alpha < \max(\min(p_t(\pi_a), p_t(\pi_b)), \epsilon)$. We further assume that $p_t(\pi_a) > 0$, which is naturally satisfied by many no-regret strategies that play in the strict interior of the simplex. With these choices, plugging on our preceding setting for $p$, and dividing both sides by $\alpha$, we get that 
\begin{align*}
    \nabla R_i(p_t(\pi_a)) + \eta L_t(\pi_a) - \nabla R_i(p_t(\pi_b)) - \eta L_t(\pi_b) \geq 0.
\end{align*}
Rearranging terms and combining with our second property, we get that 
\begin{align}
    \nabla R_i(p_t(\pi_b)) - \nabla R_i(p_t(\pi_a))\leq \eta(L_t(\pi_a) - L_t(\pi_b)) \leq -\eta \sum_{s=1}^{t-1} q_s(\Pi^\star) \Delta/2. \label{eq:star}
\end{align}
We now specialize to the case of Hedge, where $R_i(x) = x\ln x$ and $\nabla R_i(x) = \ln x + 1$ but note that a similar argument holds for a wide variety regularizers (i.e. other no-regret algorithms) under analogous assumptions. Then, by simplifying the RHS and LHS of our preceding expression and exponentiating both sides, we have that
\begin{align*}
    p_t(\pi_b) \leq p_t(\pi_a) \exp\left(\frac{-\eta\Delta}{2} \sum_{s=1}^{t-1} q_s(\Pi^\star)\right).
\end{align*}
Because the term inside the exponential is always negative and therefore the exponential is at most 1, the above expression implies that $\forall \pi_b \notin \Pi^\star$, $p_t(\pi_b) \leq p_t(\pi_a) \leq p_t(\Pi^\star)$. Thus, via Hölder's inequality, we have that $1 = \sum_{\pi \in \Pi} p_t(\pi) \leq |\Pi| p_t(\Pi^\star) \Rightarrow \frac{1}{|\Pi|} \leq p_t(\Pi^\star) \Rightarrow \exp(-p_t(\Pi^\star)) \leq \exp(\frac{-1}{|\Pi|})$. Since the same holds for $q_t$ by symmetry, we can conclude that 
\begin{align*}
    p_t(\pi_b) \leq p_t(\pi_a) \exp\left( \frac{-\eta t\Delta}{2|\Pi|} \right).
\end{align*}

Consequently, once we have $\exp(-\eta t\Delta/(2|\Pi|)) \leq \epsilon$, the probability of any action $b \notin \Pi^\star$ is at most $\epsilon$. Inverting the preceding expression shows this happens in at most $t \leq \frac{2|\Pi|}{\eta\Delta} \ln \frac{1}{\epsilon}$ rounds. Hence,
\begin{align*}
    \mathsf{Reg}(T) \leq \epsilon T + \frac{2|\Pi|}{\eta\Delta} \ln \frac{1}{\epsilon}. 
\end{align*}
Choosing $\eta = 1$ and $\epsilon = 1/T$ gives a fast rate. 
\end{pf}

We note that the linear dependence on $\Pi$ is similar to the linear dependence on the number of actions in most gap-dependent bounds for UCB algorithms. We also note that in the bandit feedback setting, Equation \ref{eq:star} only changes in that we have importance-sampled estimates $\hat L_t(a) - \hat L_t(b)$ instead of the population losses. Adding and subtracting the true means gives us the same quantity as Eq. \ref{eq:star} plus a martingale. This term is $O(\eta \sqrt{t})$. For $\sqrt{t} = O(t\Delta/|\Pi|)$, or equivalently, $t = O((|\Pi|/\Delta)^2)$, we get the same bound with slightly different constants.

\subsection{Extension to Bandit Feedback}
\label{app:proofs:bandits}
We now provide an extension of our main result to the bandit feedback setting via the standard importance sampling argument.

\begin{theorem} Assume $\Pi$ is finite. Let $\pi_t, \pi'_t \sim p_t$. For any fixed $\alpha \in [0, 1]$ and for some $\gamma \in [0, 1]$, if we set $\mathcal{O}$ to be the Hedge algorithm of \citet{freund1997decision} and feed it the sequence of loss functions  \begin{equation}\hat{\ell}_t^{\texttt{SPO}}(\pi)  = -\alpha \frac{\mathbbm{1}[\pi = \pi_t]}{p_t(\pi_t)} \mathcal{P}(\pi_t, \pi_t') + (1 - \alpha)\frac{\mathbbm{1}[\pi = \pi_t']}{p_t(\pi_t')} \mathcal{P}(\pi_t, \pi_t'), \nonumber \end{equation} and follow strategy $(1 - \gamma) p_t + \frac{\gamma}{|\Pi|}$, $\bar{p}$ is an $\widetilde{O}(\frac{1}{\sqrt{T}})$-approximate Minimax Winner. \footnote{For last iterate convergence, one can instead run the no-regret bandit feedback algorithm of \citet{cai2023uncoupled} on $\hat{\ell}_t^{\texttt{SPO}}$, albeit at the cost of a slower rate.}
\label{corr:spo_bf}
\end{theorem}
\begin{pf} Let $\pi_t \sim p_t$, $\pi_t' \sim q_t$. We observe $\mathcal{P}(\pi_t, \pi_t')$ by querying the preference function. 

Define 
\begin{equation}
    \hat{\ell}_t^1(\pi) = -\frac{\mathbbm{1}[\pi = \pi_t]}{p_t(\pi_t)} \mathcal{P}(\pi_t, \pi_t'), \quad \hat{\ell}_t^2(\pi) = \frac{\mathbbm{1}[\pi = \pi_t']}{q_t(\pi_t')} \mathcal{P}(\pi_t, \pi_t')
\end{equation}
and as the importance-weighted losses for each player at round $t$. Next, observe that
\begin{equation}
    \mathbb{E}_{p_t}[\hat{\ell}_t^1] = \ell_t^1, \mathbb{E}_{q_t}[\hat{\ell}_t^2] = \ell_t^2,
\end{equation}
i.e. that they are unbiased estimates of the full-feedback losses. We now proceed similarly to our proof for the full feedback case. If we set $p_0 = q_0$, we have that $\ell_0^1 = \ell_0^2$ by the anti-symmetry of $\mathcal{P}$. We also have that 
\begin{equation}
   \mathbb{E}_{p_0}[\hat{\ell}_0^{\texttt{SPO}}] =  \mathbb{E}_{p_0}[ \alpha \hat{\ell}_0^1 + (1 - \alpha) \hat{\ell}_0^2] = \ell_0^1 = \ell_0^2 .
\end{equation}
We now proceed by induction. If $p_{\tau} = q_{\tau}$ at some time $\tau$ and we feed both players the same loss $\hat{\ell}_{\tau}^{\texttt{SPO}}$, by the determinism of Hedge, we have that $p_{\tau + 1} = q_{\tau + 1}$. This implies $\ell_{\tau+1}^1 = \ell_{\tau + 1}^2$, which in turn implies $\hat{\ell}_{\tau + 1}^{\texttt{SPO}}$ is an unbiased estimate of both $\ell_{\tau + 1}^1$ and $\ell_{\tau + 1}^2$. Thus, by induction, we have that both players will have the same iterates and $\hat{\ell}_t^{\texttt{SPO}}$ is an unbiased estimate of both losses $\forall t \in [T]$, which means we can simulate game-solving with a single player.

To prove that above loss estimation scheme preserves the no-regret property, we can simply examine the proof of Exp3 in \citet{auer2002nonstochastic} and note that the only property required of the loss estimate ($\hat{x}_t$ in the original paper) is that it is unbiased. Thus, we inherit the regret rate of Exp3, which when plugged in completes the proof.

\end{pf}
We note that while the above is not a general reduction per se, for a wide set of no-regret algorithms, a similar argument applies, with slight differences in the effect of importance sampling on the final regret rate.

\subsection{Extension to Contextual Setting}
\label{app:proofs:context}
We now extend our above setup to include contexts. Consider a finite-horizon reward-free contextual Markov Decision Process (MDP) \citep{puterman2014markov} parameterized by $\langle \mathcal{S}, \mathcal{A}, \mathcal{X}, \mathcal{T}, H , \rho \rangle$ where $\mathcal{S}$, $\mathcal{A}$, $\mathcal{X}$ are the state, action, and context spaces, $\mathcal{T}: \mathcal{S} \times \mathcal{A}\rightarrow \Delta(\mathcal{S})$ is the transition operator, $H$ is the horizon, and $\rho : \Delta(\mathcal{X})$ is the context / initial state distribution.  We use $\Xi \triangleq (\mathcal{S} \times \mathcal{A}) ^H$ to denote the space of trajectories and $\Phi_h \triangleq \mathcal{X} \times (\mathcal{S} \times \mathcal{A}) ^{h-1} \times \mathcal{S}$ to denote the space of $h$-length histories. We assume that we are given access to a \textit{(contextual) preference function}\begin{equation}
    \mathcal{P}: \mathcal{X} \times \Xi \times \Xi \rightarrow [-1, 1] 
\end{equation}
which, given two trajectories $\xi_1, \xi_2 \in \Xi$, outputs a scalar that indicates which is preferred relative to the other. By construction, preference functions are anti-symmetric, i.e. $\forall x, \xi_1, \xi_2 \in \mathcal{X} \times \Xi \times \Xi $, $\mathcal{P}(x, \xi_1, \xi_2) = -\mathcal{P}(x, \xi_2, \xi_1)$. Similarly, we also have that $\forall x, \xi \in \mathcal{X} \times \Xi$, $ \mathcal{P}(x, \xi, \xi) = 0$. We assume access to a convex and compact policy class $\Pi \subseteq \{\mathcal{S} \times \mathcal{X} \rightarrow \Delta(\mathcal{A})\}$. 
With a slight abuse of notation, we can define the preference function over policy pairs as
\begin{equation}
    \mathcal{P}(\pi_1, \pi_2) \triangleq \mathbb{E}_{x \sim \rho, \xi_1 \sim (\pi_1, x), \xi_2 \sim (\pi_2, x) }[\mathcal{P}(x, \xi_1, \xi_2)]. \nonumber
\end{equation}

We now re-state our algorithms, including the dependence on context. Their theoretical guarantees match those presented in the main paper. The main difference in practice is that rather than simply maintaining a queue, we now have to sample multiple trajectories based on a single context for comparison.
\begin{algorithm}[h]
\begin{algorithmic}
\STATE {\bfseries Input:} Learning rate $\eta$, Iterations $T$, Preference fn. $\mathcal{P}$.
\STATE {\bfseries Output:} Trained policy $\pi$.
\STATE Initialize $\pi_h(\cdot \vert \phi) = \mathsf{Unif}(\mathcal{A})$, $\forall \phi \in \Phi_h, h \in [H]$.
\FOR{$t$ in $1 \dots T$}
\STATE Observe context $x_t \sim \rho$.
\STATE Compute $r_t(\xi) = \mathbb{E}_{\xi' \sim \pi_t}\mathcal{P}(x_t, \xi, \xi')$.
\STATE Define $Q_t^h(\phi, a) = \mathbb{E}_{\xi \sim \pi_t}[r_t(\xi)|\phi_h = \phi, a_h = a]$.
\STATE Define $A_t^h(\phi, a) = Q_t^h(\phi, a) - \mathbb{E}_{a' \sim \pi_t^h(\phi)}[Q_t^h(\phi, a')]$.
\FOR{$\phi, a, h \in \Phi_h \times \mathcal{A} \times [H]$}
\STATE \algcommentlight{use no-regret algo for update}
\STATE $\pi_{t+1}^h(a \vert \phi) \propto \pi_{t}^h(a \vert \phi) \exp{(\eta A_t^h(\phi, a))}$.
\ENDFOR
\ENDFOR
\STATE {\bfseries Return } $\bar{\pi}$, uniform mixture of $\pi_{1:T}$.
\end{algorithmic}
\caption{\texttt{SPO} (Theoretical Version with Contexts) \label{alg:spo-theory-contexts}}
\end{algorithm}

\begin{algorithm}[h]
\begin{algorithmic}
\STATE {\bfseries Input:} Iterations $T$, Preference fn. $\mathcal{P}$, Num. samples $k \geq 2$, Reinforcement learning algo. $\mathsf{RL}: \Pi \times \mathcal{D} \to \Pi$.
\STATE {\bfseries Output:} Trained policy $\pi$.
\STATE Initialize $\pi_1 \in \Pi$.
\FOR{$t$ in $1 \dots T$}
\STATE Observe context $x_t \sim \rho$ and sample $\xi_{1:k} \sim \pi_t(x_t)$.
\STATE Compute $r_t(\xi_i) = \frac{1}{k-1}\sum_{j \neq i}^k \mathcal{P}(x_t, \xi_i, \xi_j)$.
\STATE Set $r_i^h = r_t(\xi_i) / H$, $\forall i, h \in [k] \times [H]$.
\STATE \algcommentlight{use PPO, TRPO, SAC ...}
\STATE $\pi_{t+1} \gets \mathsf{RL}(\pi_t, \mathcal{D} = \{(s^h_i, a^h_i, r_i^h)\}_{i \in [k]}^{h \in [H]})$. 
\ENDFOR
\STATE {\bfseries Return } best of $\pi_{1:T}$ on validation data.
\end{algorithmic}
\caption{\texttt{SPO} (Practical Version with Contexts) \label{alg:spo-praxis-contexts}}
\end{algorithm}

\clearpage
\section{Compounding Errors in RLHF}
\label{app:compounding_errs}
A common concern in sequential prediction tasks is \textit{compounding errors} \citep{ross2011reduction}. Consider, for example, trying to train a policy to drive laps around a track purely based on recorded demonstrations from an expert driver who always stays close to the center of their lane. For example, one could regress from recorded states to recorded actions, an approach known as behavioral cloning (BC, \citet{pomerleau1988alvinn}). However, if at test time, the agent makes a mistake and goes off the center of their lane, they might end up in a state they hadn't seen in their training data, have no idea what to do in this novel situation, make another error, and quickly spiral out of control. At a fundamental level, the agent's poor test-time performance is caused by the \textit{covariate shift} in terms of state distribution between the offline training data and their own \textit{induced} state distribution -- a low training error does not necessarily correspond to a low test error. Thus, the standard solution is to allow the learner to actually try out actions in the environment, see where they end up, and learn to correct their mistakes. This approach is known as inverse reinforcement learning (IRL, \citet{ziebart2010modeling}) and is known to prevent compounding errors \citep{swamy2021moments}.

A natural question might be if the learner gets preferences rather than demonstrations as feedback, whether the same concerns arise. We now provide a simple, informal example of this.
\begin{example}
Consider the $H=2$ problem of completing sentences of the form ``\textsc{The (a) orbits around the (b).}'' where either blank can be filled in with utterances $\mathcal{A}$ = \{\textsc{earth, sun, moon}\}. We observe preferences of the form [\textsc{moon, earth}] $\succ$ [\textsc{moon, sun}]. At $h = 1$, we are forced to play a policy $\pi_{\epsilon}$ that outputs \textsc{moon} w.p. $1 - \epsilon$ and \textsc{earth} w.p. $\epsilon$. At $h=2$, we are allowed to choose between two policies: $\pi_1$ that always outputs \textsc{earth} and $\pi_2$ that outputs \textsc{earth} if the preceding word was \textsc{moon} and \textsc{sun} if the preceding word was \textsc{earth}. Observe that both $[\pi_{\epsilon}, \pi_1]$ and $[\pi_{\epsilon}, \pi_2]$ have the same probability of generating the preferred and dis-preferred generations and therefore will have the same value under \textbf{any} loss function that depends on its inputs purely via their off-policy likelihoods.
\label{ex:orbits}
\end{example}
We can think of $\pi_{\epsilon}$ as representing error from finite samples, imperfect optimization, or a limited function class. While hopefully any post-Copernican preference (or reward) model would be able to tell the difference between these two policies, offline approaches like DPO that simply compute likelihoods on off-policy data are unable to do so.  \footnote{An iterative application of DPO with batches of preferences collected frequently would likely mitigate this issue.} We emphasize that this is a fundamental issue with \textit{all} offline approaches, rather than with a particular offline algorithm \citep{swamy2021moments}. We conclude with a note that this problem only gets worse with longer task horizons as there are more timesteps to deviate. This is perhaps one of the reasons that offline approaches can sometimes under-perform interactive techniques \citep{starling2023, chen2024selfplay, tajwar2024preference, xu2024dpo, tang2024understanding, song2024understanding} and why, as of the writing of this paper, the world's most performant models are interactively trained \citep{team2023gemini, openai2023gpt4}. 

\clearpage
\section{Additional Results}
\label{app:extra_res}
\begin{figure*}[h]
    \centering
    \includegraphics[width=0.4\textwidth]{figures/max_reward_pref_cheetah_run.pdf}
    \includegraphics[width=0.4\textwidth]{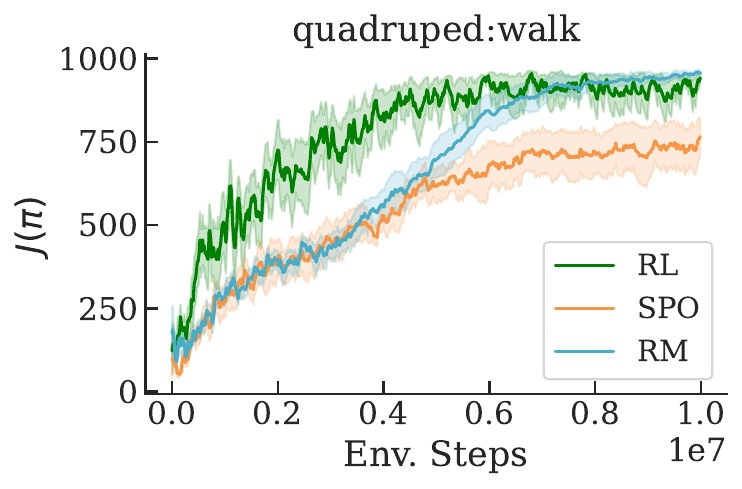} 
    \includegraphics[width=0.4\textwidth]{figures/max_reward_pref_quadruped_run.pdf}
    \includegraphics[width=0.4\textwidth]{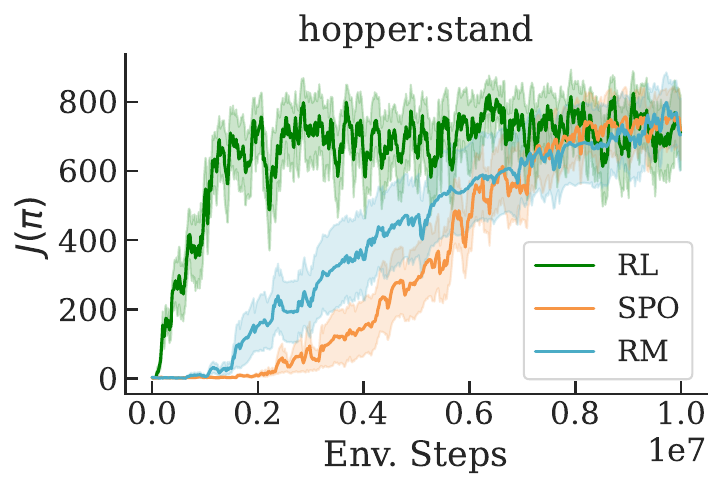}
    \includegraphics[width=0.4\textwidth]{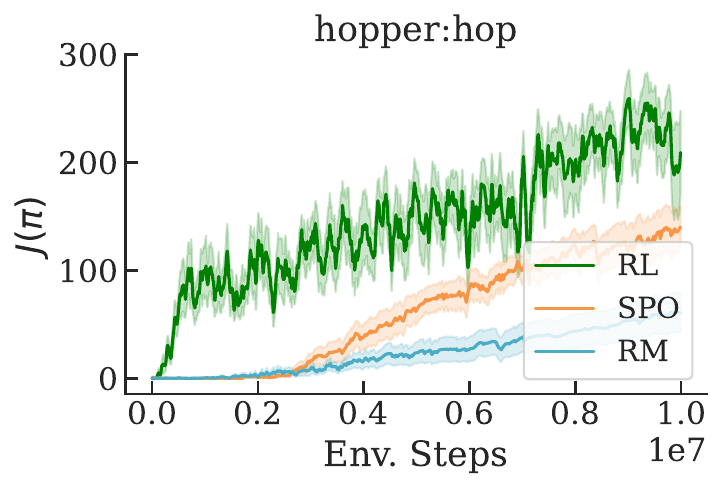}
    \includegraphics[width=0.4\textwidth]{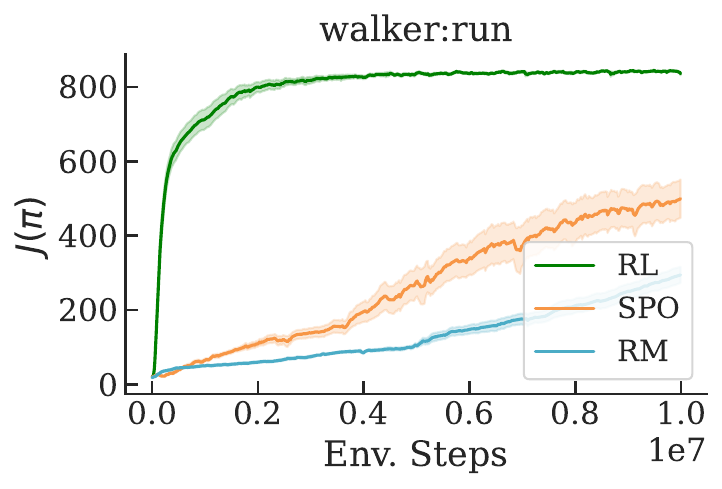}
    \includegraphics[width=0.4\textwidth]{figures/max_reward_pref_walker_stand.pdf}
    \includegraphics[width=0.4\textwidth]{figures/max_reward_pref_walker_walk.pdf}
    \caption{All results for maximum reward preference experiments.}
    \label{fig:max_reward_extra}
\end{figure*}

\begin{figure*}[h]
    \centering
    \includegraphics[width=0.4\textwidth]{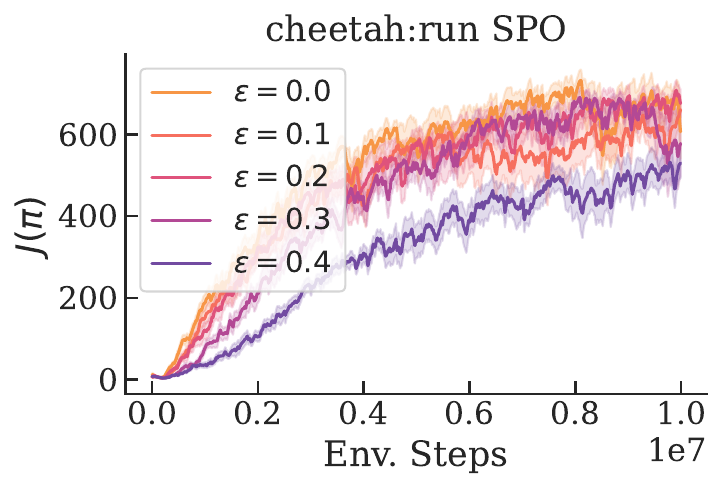}
    \includegraphics[width=0.4\textwidth]{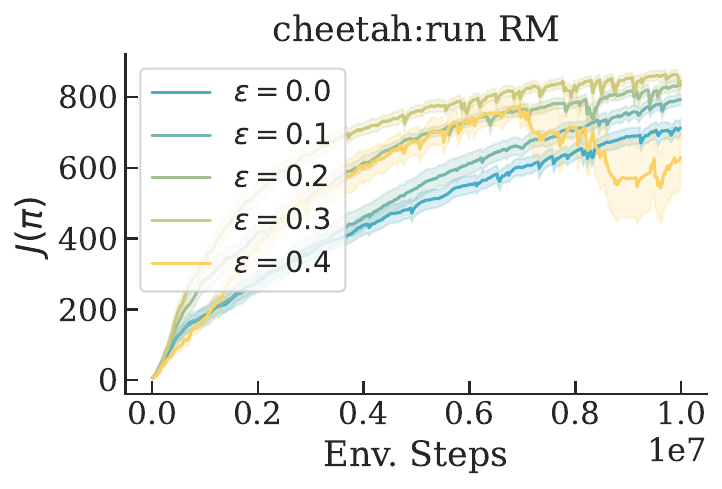}
    \includegraphics[width=0.4\textwidth]{figures/noisy_pref_spo_walker_stand.pdf}
    \includegraphics[width=0.4\textwidth]{figures/noisy_pref_rm_walker_stand.pdf}
    \includegraphics[width=0.4\textwidth]{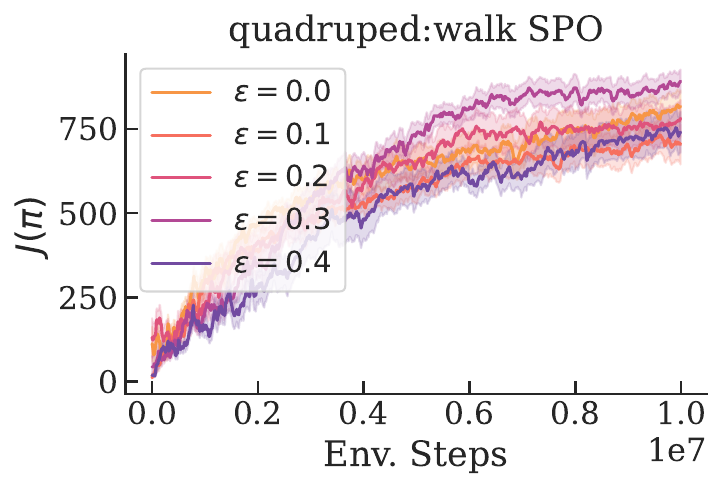}
    \includegraphics[width=0.4\textwidth]{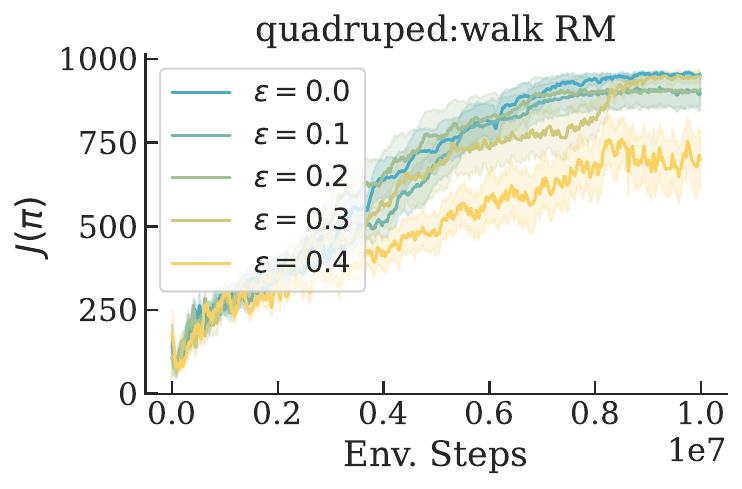}
    \includegraphics[width=0.4\textwidth]{figures/noisy_pref_spo_walker_walk.pdf}
    \includegraphics[width=0.4\textwidth]{figures/noisy_pref_rm_walker_walk.pdf}
    \caption{All results for noisy preference experiments.}
    \label{fig:noisy_feedback_extra}
\end{figure*}

\begin{figure*}[h]
    \centering
    \includegraphics[width=0.4\textwidth]{figures/nonmarkov_pref_quadruped_run_50.0.pdf}
    \includegraphics[width=0.4\textwidth]{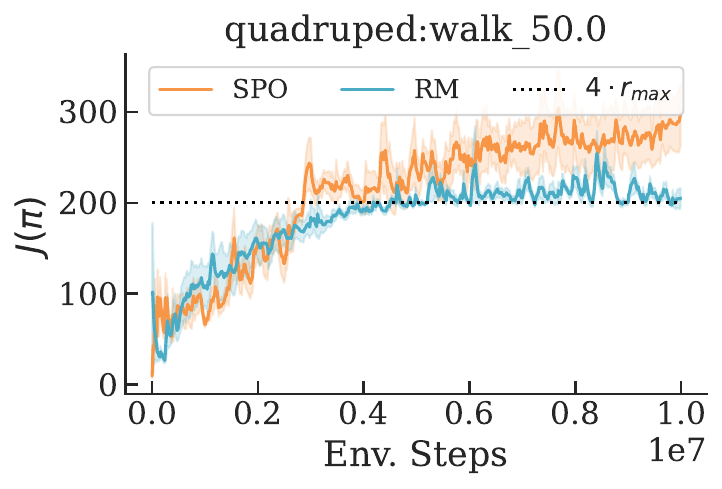}
    \includegraphics[width=0.4\textwidth]{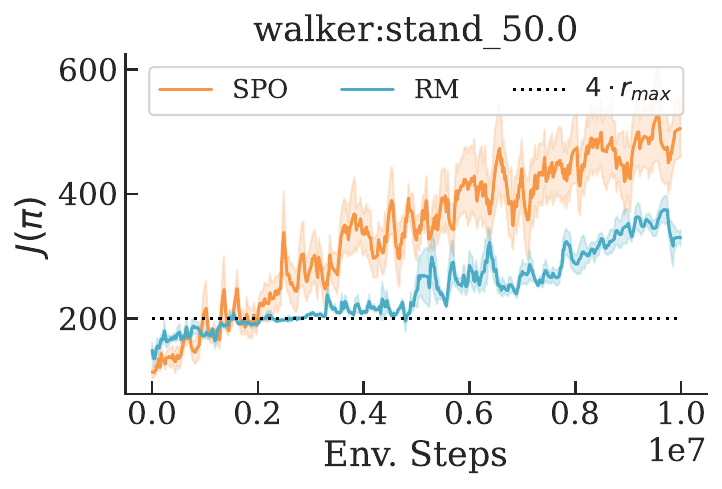}
    \includegraphics[width=0.4\textwidth]{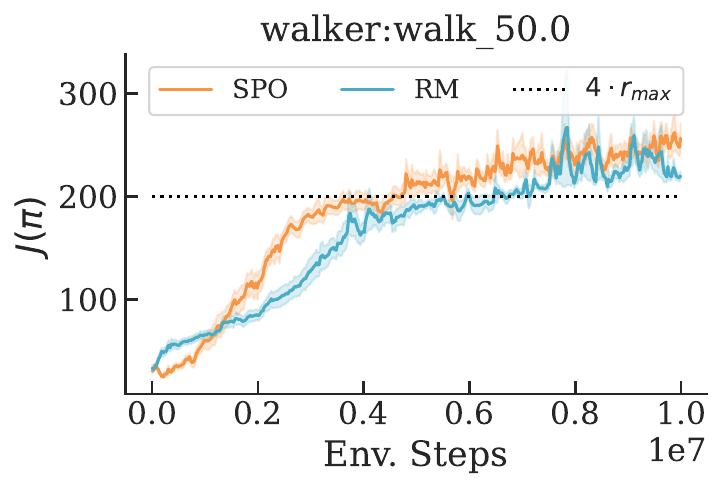}
    \includegraphics[width=0.4\textwidth]{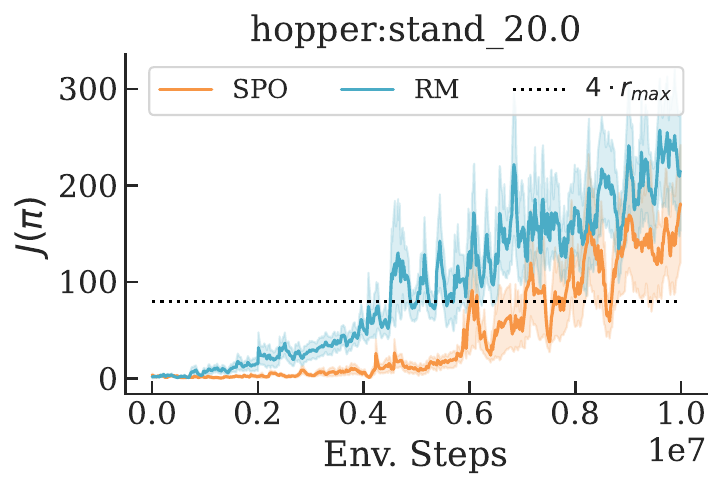}
    \includegraphics[width=0.4\textwidth]{figures/nonmarkov_pref_cheetah_run_10.0.pdf}
    \caption{All results for non-markovian preference experiments.}
    \label{fig:nonmarkov_extra}
\end{figure*}

\begin{figure*}[h]
        \centering
        \includegraphics[width=0.8\textwidth]{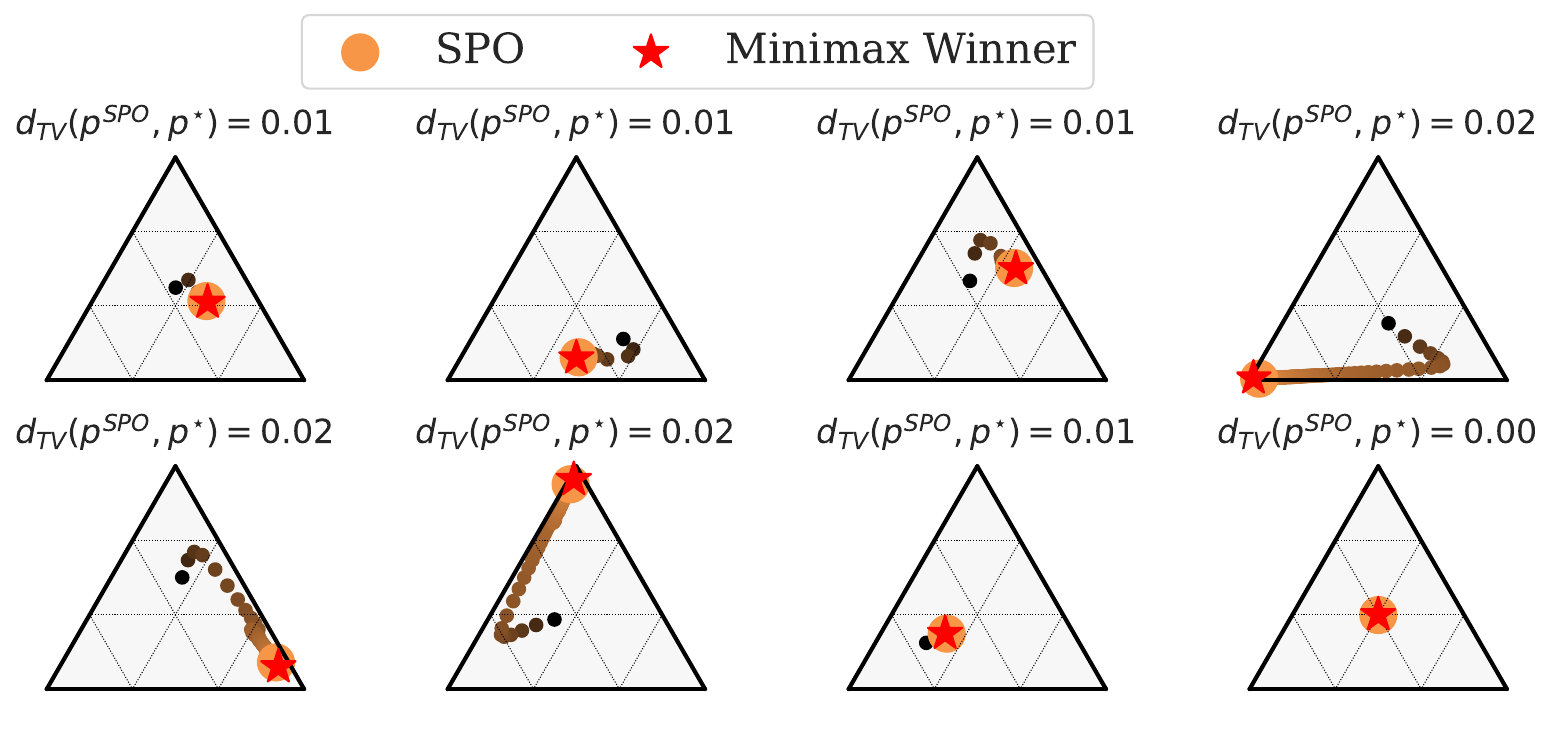}
        \caption{Convergence of SPO on bandit instances with non-transitive preferences.}
    \label{fig:rps_bandit_spo}
\end{figure*}

\begin{figure*}[h]
        \centering
        \includegraphics[width=0.8\textwidth]{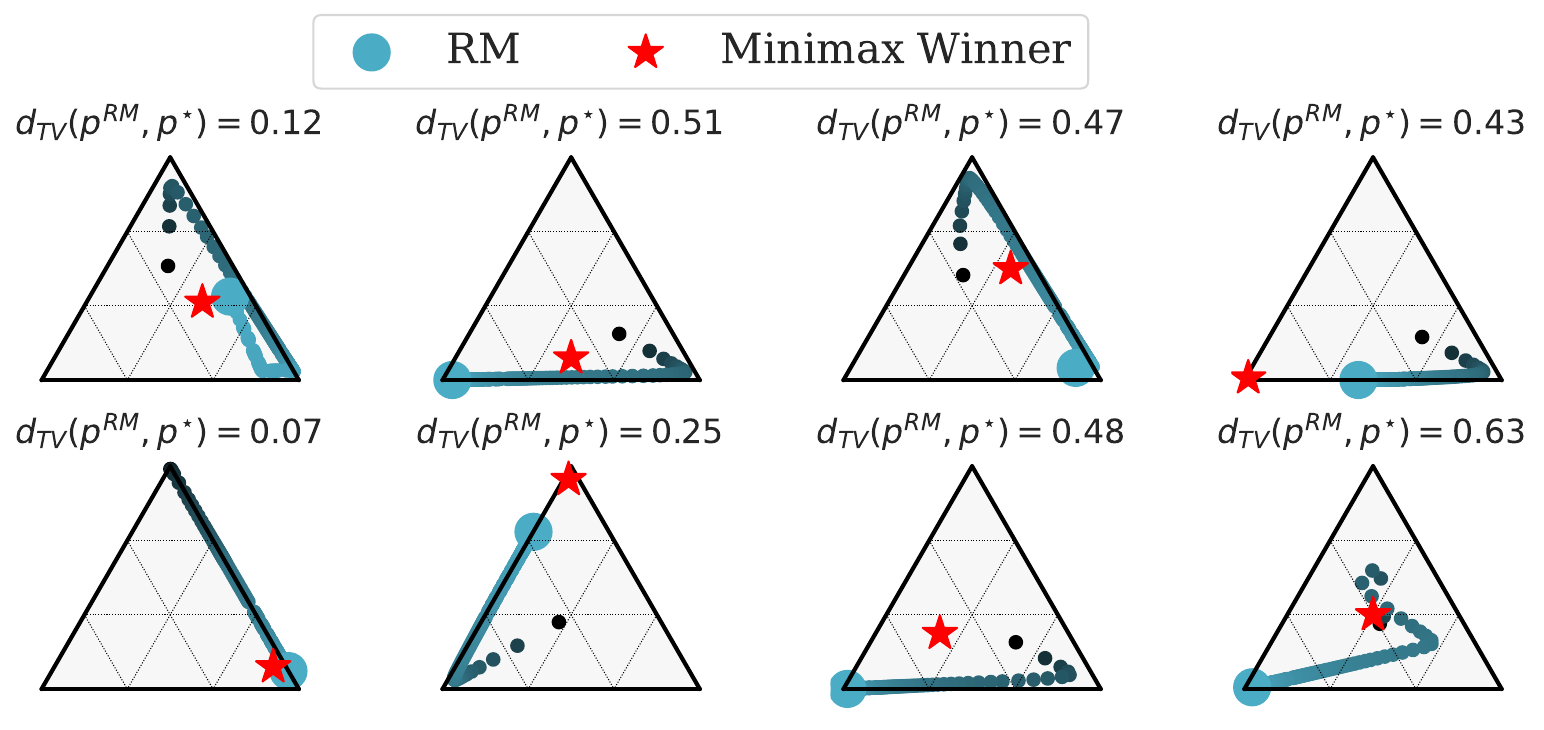}
        \caption{Convergence of RM on bandit instances with non-transitive preferences.}
    \label{fig:rps_bandit_rm}
\end{figure*}

\begin{figure*}[h]
    \centering
    \begin{subfigure}{0.49\textwidth}
    \centering
    \includegraphics[width=0.35\textwidth]{figures/ant_cyclic_0.pdf}
    \includegraphics[width=0.6\textwidth]{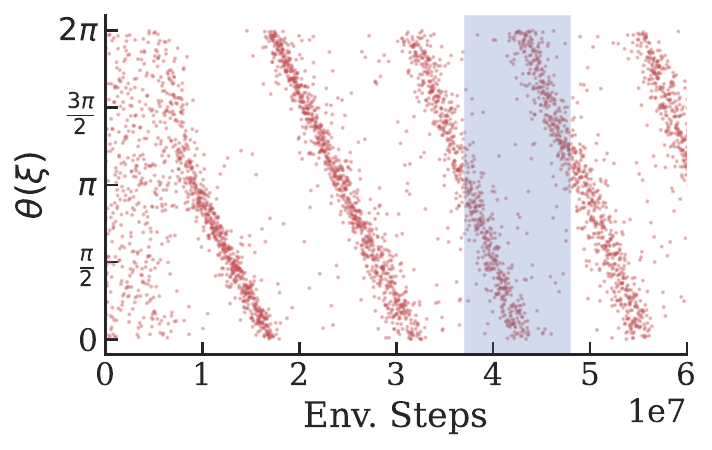}
    \caption{Seed 1.}
    \end{subfigure}
    \centering
    \begin{subfigure}{0.49\textwidth}
    \centering
    \includegraphics[width=0.35\textwidth]{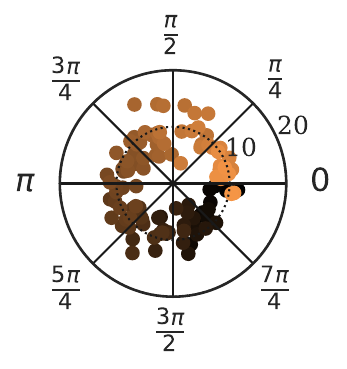}
    \includegraphics[width=0.6\textwidth]{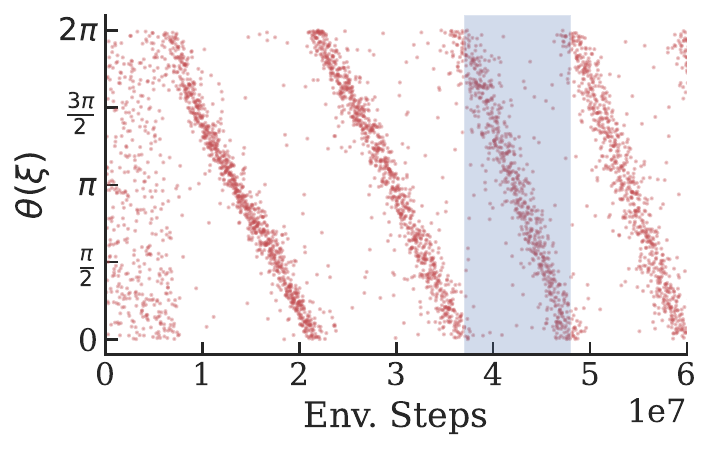}
    \caption{Seed 2.}
    \end{subfigure}
    \centering
    \begin{subfigure}{0.49\textwidth}
    \centering
    \includegraphics[width=0.35\textwidth]{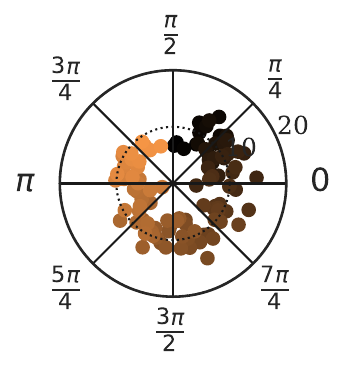}
    \includegraphics[width=0.6\textwidth]{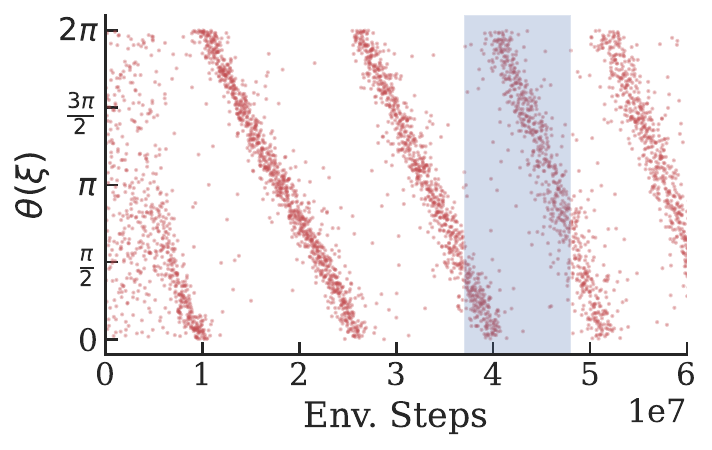}
    \caption{Seed 3.}
    \end{subfigure}
    \centering
    \begin{subfigure}{0.49\textwidth}
    \centering
    \includegraphics[width=0.35\textwidth]{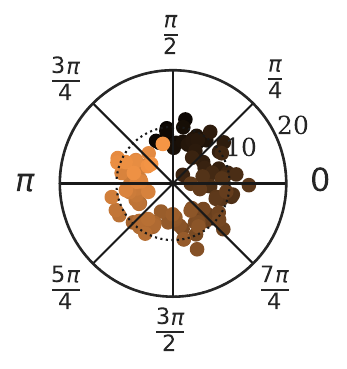}
    \includegraphics[width=0.6\textwidth]{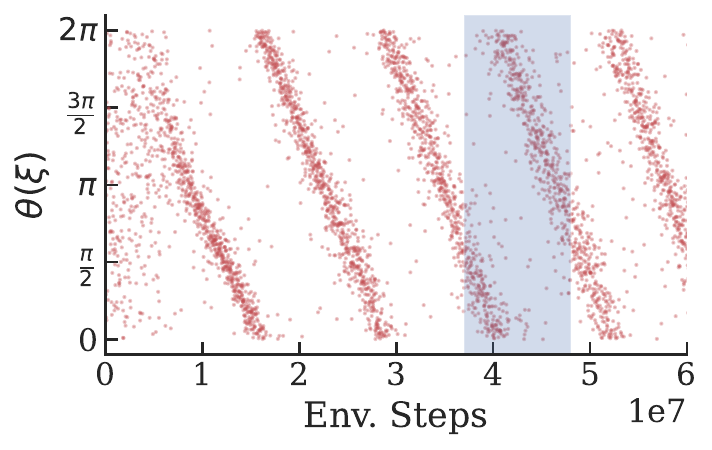}
    \caption{Seed 4.}
    \end{subfigure}
    \caption{Additional results for intransitive preference experiments with Gym Ant-v3 environment. Here, we present results with $4$ seeds with these trends holding across all the $10$ seeds we experimented with. As presented in the main paper, in the polar plots, each dot corresponds to 100k timesteps of training (where the darker shade represents earlier environment steps and lighter shade represent later environment steps). We see our agent traces out a circle of radius 10 (qualitatively the behavior we expect from MW) on average. Alongside the polar plots, we plot the agent's angle through the first 60 million steps of training with the shaded region representing the cross section of iterations where the agent's position and angle are plotted in the corresponding polar plot.}
    \label{fig:cyclic_extra}
\end{figure*}

\begin{figure*}[h]
    \centering
    \includegraphics[width=0.4\textwidth]{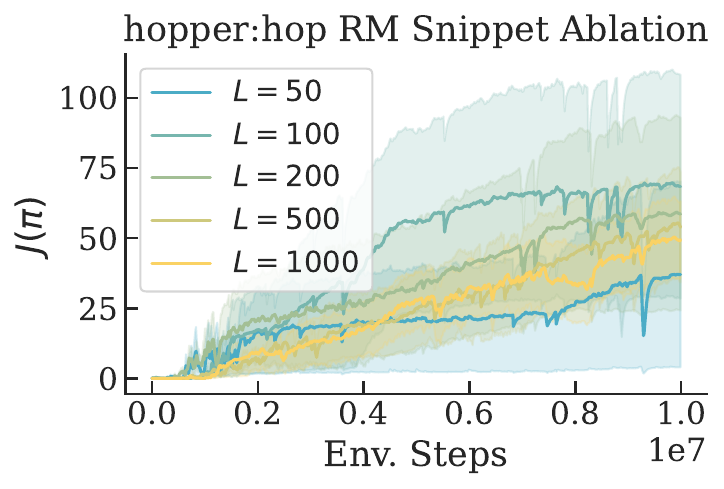}
    \includegraphics[width=0.4\textwidth]{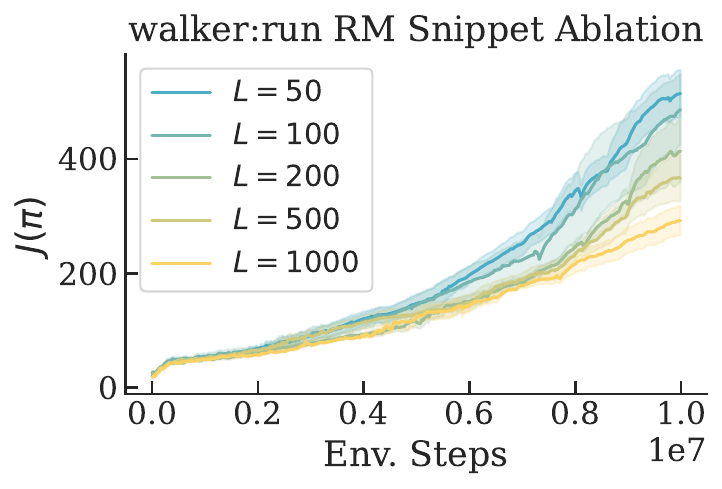}
    \includegraphics[width=0.4\textwidth]{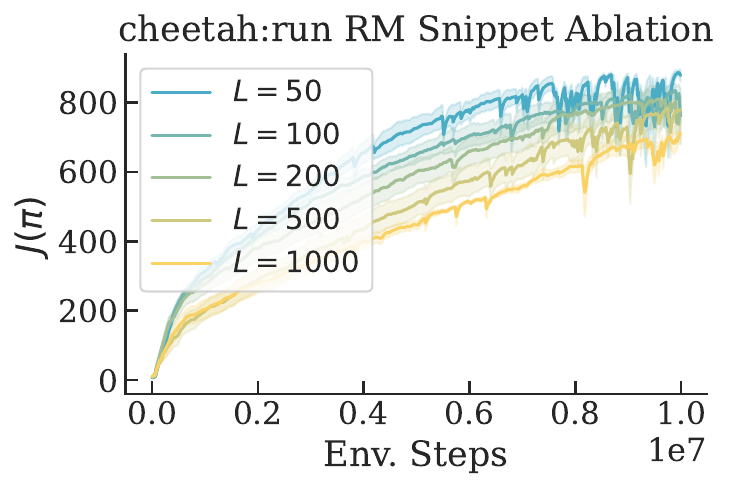}
    \includegraphics[width=0.4\textwidth]{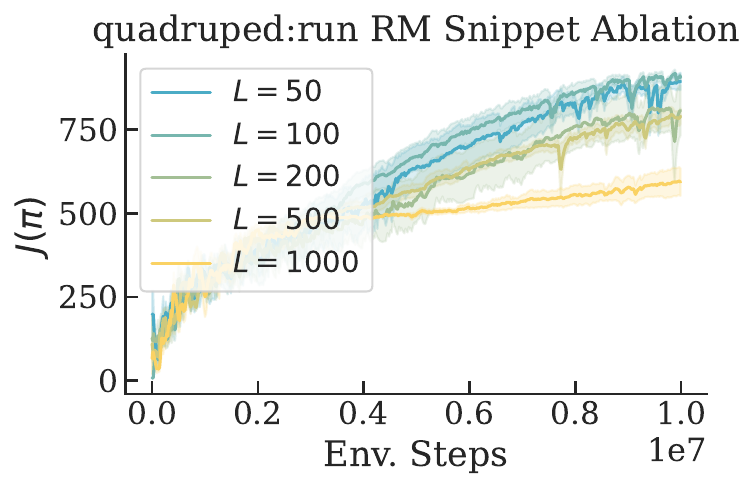}
    \includegraphics[width=0.4\textwidth]{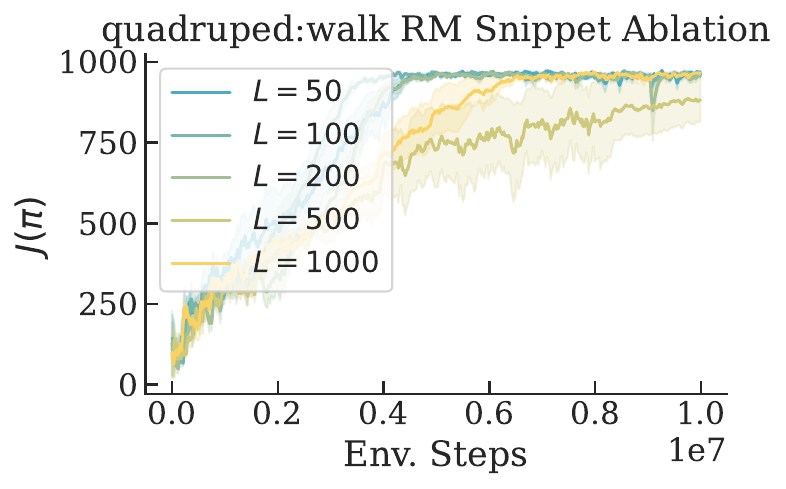}
    \includegraphics[width=0.4\textwidth]{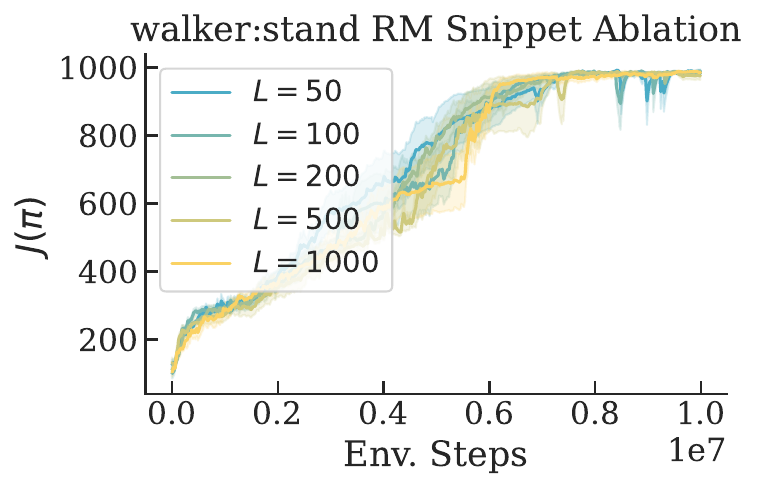}
    \includegraphics[width=0.4\textwidth]{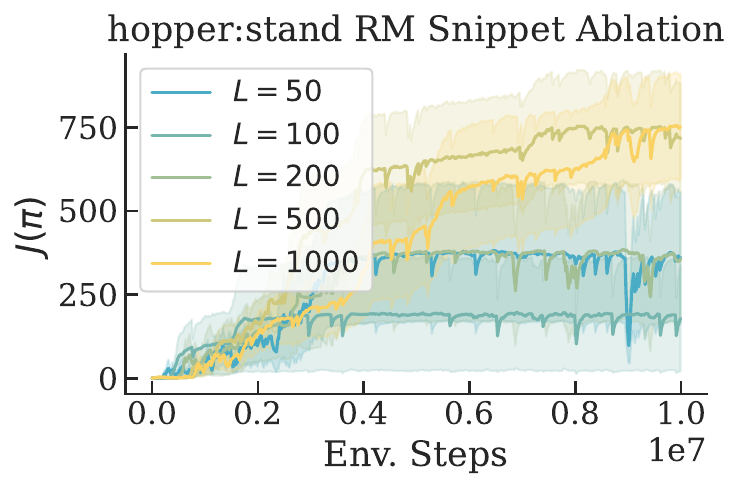}
    \includegraphics[width=0.4\textwidth]{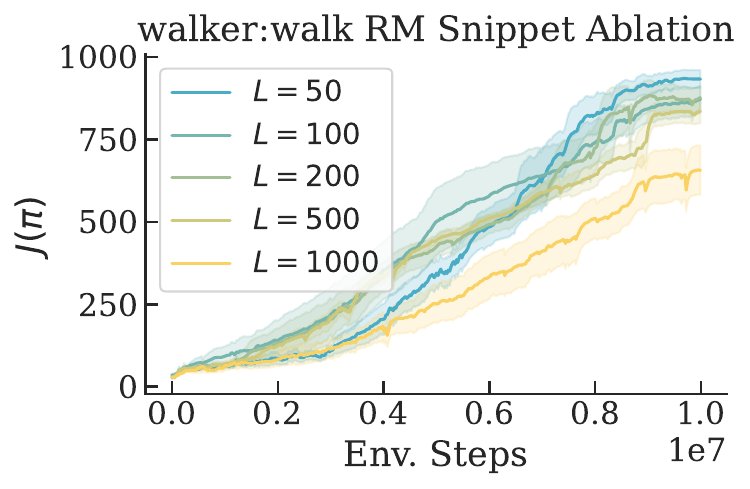}
    \caption{All results for reward model snippet length ablations.}
    \label{fig:snippet_ablation}
\end{figure*}

\begin{figure*}[h]
    \centering
    \includegraphics[width=0.4\textwidth]{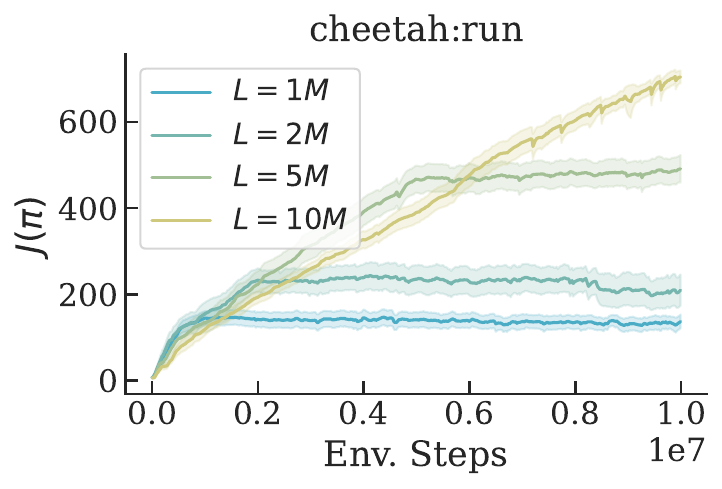}
    \includegraphics[width=0.4\textwidth]{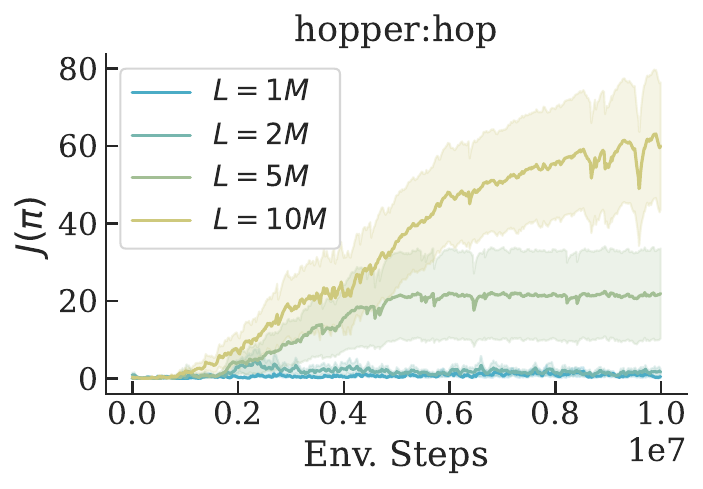}
    \includegraphics[width=0.4\textwidth]{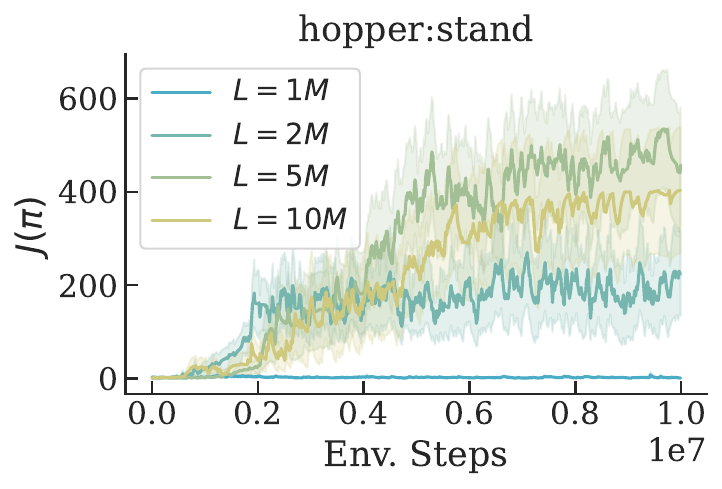}
    \includegraphics[width=0.4\textwidth]{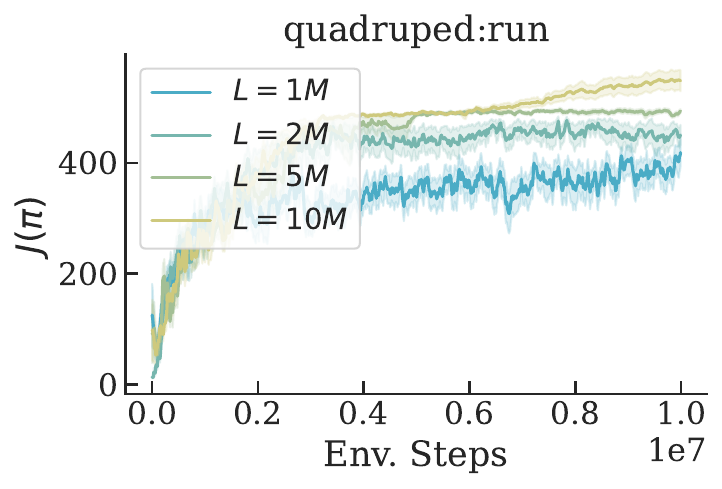}
    \includegraphics[width=0.4\textwidth]{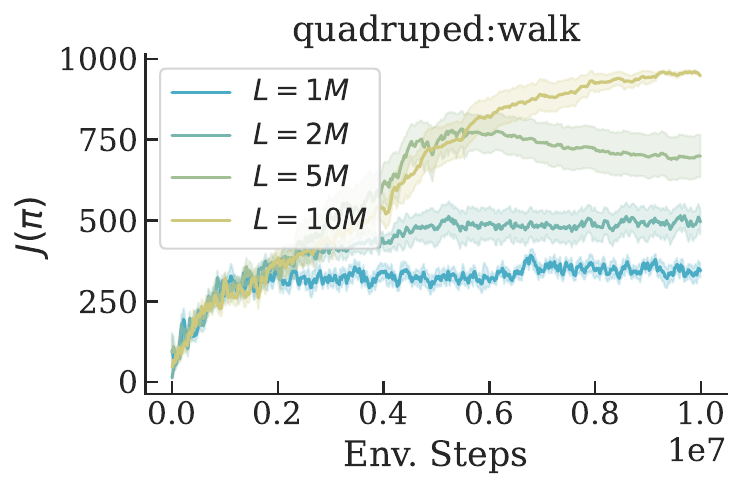}
    \includegraphics[width=0.4\textwidth]{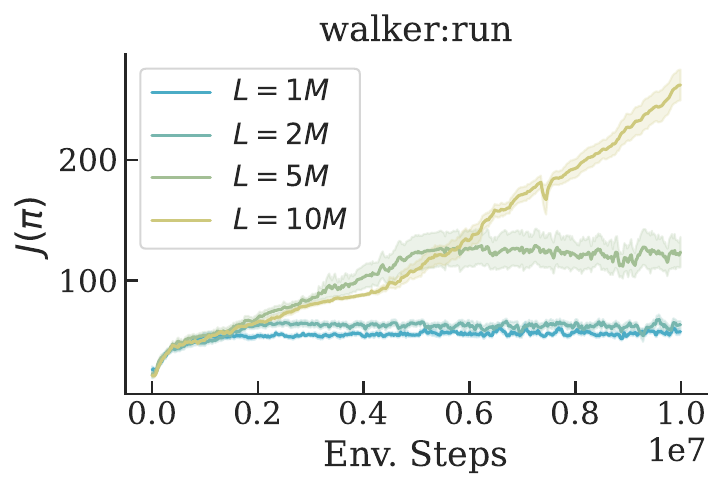}
    \includegraphics[width=0.4\textwidth]{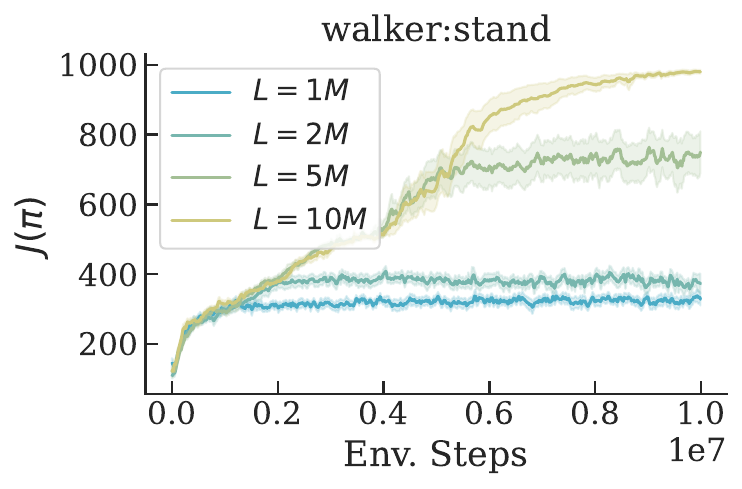}
    \includegraphics[width=0.4\textwidth]{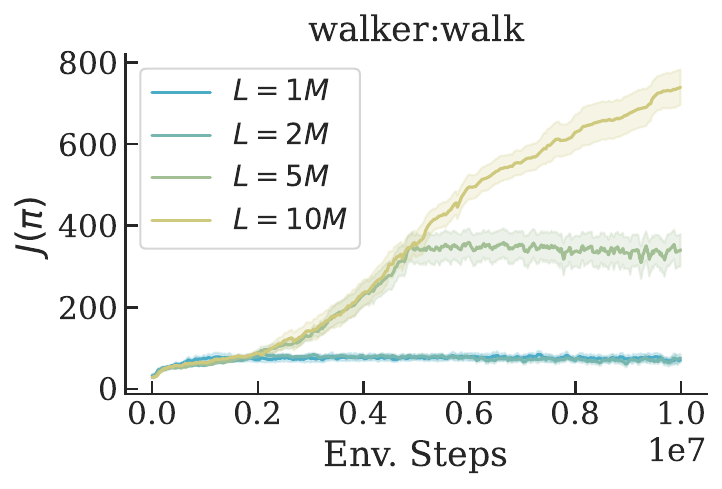}
    \caption{All results for reward model freezing experiments.}
    \label{fig:frozen_rm}
\end{figure*}

We now present additional results we did not have space for in the main paper:
\begin{itemize}
\item Figure~\ref{fig:max_reward_extra} contains results on a larger set of control environments with preferences based on reward maximization. 
\item Figure~\ref{fig:noisy_feedback_extra} contains results on a larger set of control environments with noisy preferences based on reward maximization. 
\item Figure~\ref{fig:nonmarkov_extra} contains results on a larger set of control environments with non-Markovian preferences. 
\item Figure~\ref{fig:rps_bandit_spo} contains results for SPO on a larger set of bandit environments with intransitive preferences. Figure~\ref{fig:rps_bandit_rm} contains the results for RM in the same setting.
\item Figure~\ref{fig:cyclic_extra} shows additional seeds for SPO the Ant environment with intransitive preferences.
\item Figure~\ref{fig:snippet_ablation} ablates the effect of different snippet lengths on RM sample efficiency.
\item Figure~\ref{fig:frozen_rm} contains the effect of freezing the reward model during training on RM performance.
\end{itemize}

\paragraph{Number of queries to the preference oracle.} We report the performance of an agent as a function of the total steps in the environment steps. We did not attempt to optimize performance of SPO or RM as a function of the number of calls to the preference oracle it makes. That said, in our experiments, SPO makes fewer oracle calls than RM. The total number of calls to an oracle made by SPO are the number of episodes collected times the queue length $B$, since each generated episode is compared against the previous $B$ ones. In comparison, the total number of calls made by RM are the number of updates to the reward model times the batch size.
For example, for the experiment with maximum-reward preferences (Figure~\ref{fig:max_reward_extra}), this is a total of \textbf{1M calls for SPO} and \textbf{2.5M calls for RM}.



\clearpage
\section{Experimental Details}
\label{app:exps}
\localtableofcontents
\subsection{Discrete action experiment in Figure~\ref{fig:max_reward} (a)}
\label{app:exps:disc}
\paragraph{Problem Setup.}
We use a 3-armed bandit with different preference functions that each induce a different (unique) Minimax Winner. All our preference functions have the following form, represented as an $\vert \mathcal{A}\vert$ by $\vert \mathcal{A}\vert$ matrix
\begin{align*}
   \mathcal P =  \begin{bmatrix}
        0 & c & -b\\
        -c & 0 & a\\
        b & -a & 0
    \end{bmatrix}
\end{align*}
where $a, b, c > 0$ are parameters. The Minimax Winner for this preference function is $p^\star \propto \begin{bmatrix}a \\ b \\ c \end{bmatrix}$. This can be verified easily since $p^\star \mathcal{P} = \begin{bmatrix}0 \\ 0 \\ 0 \end{bmatrix}$ and thus any opponent strategy yields the same value $0$. For $a = b = c$, this is the preference in the popular Rock-Paper-Scissors game. More generally such preference functions can commonly occur when we estimate the preference from a population of users that each just have a preference between two of the three actions. Assume we have 3 sub-populations that each make up a fraction $a, b$ and $c$ of the total population, respectively. Then the average preference in the total population corresponds to $\mathcal P$
\begin{align*}
    a \times \begin{bmatrix}
        0 & 0 & 0\\
        0 & 0 & 1\\
        0 & -1 & 0
    \end{bmatrix} + b \times \begin{bmatrix}
        0 & 0 & -1\\
        0 & 0 & 0\\
        1 & 0 & 0
    \end{bmatrix} + c \times \begin{bmatrix}
        0 & 1 & 0\\
        -1 & 0 & 0\\
        0 & 0 & 0
    \end{bmatrix} = \mathcal P,
\end{align*}
where each of the 3 matrices corresponds to the preference in the respective subpopulation.

\paragraph{Common Parameters.}
We use the PPO implementation in \citet{hoffman2020acme} with learning rate $10^{-4}$. For the policy network, we use 2 hidden layer with 128 nodes each and ReLU activation. 
Since the purpose of this experiment is to verify whether a method can approach the Minimax Winner, we run each algorithm for a very large number of of steps, 50M and report the average action choice during the entire learning procedure.

\paragraph{SPO Parameters.}
We use a queue length of $B = 1000$ trajectories for computing the average preference of each trajectory. We further found that the default entropy cost of $10^{-4}$ for PPO in ACME~\citep{hoffman2020acme} to work well.

\paragraph{Reward Model Parameters.}
We use the same architecture for the reward model as for the policy network with a one-hot encoding of the action. We update the reward model every 16 policy updates, use a batch size of 64, and a learning rate of $3 \cdot 10^{-4}$. We further use an increased entropy cost of $10^{-2}$ for PPO. We found that less frequent reward-model updates and a smaller entropy cost would lead to premature convergence at arbitrary deterministic policies. Since we want to learn non-trivial MW strategies and theory suggests that a policy-dependent reward model could achieve that (see Equation~\ref{eq:pirm} and  discussion around it), We also experimented with more frequent reward-model updates. These led to instabilities in the learning process, however.

\subsection{Continuous Control Experiments}
We use the SAC implementation in \citet{hoffman2020acme} for all of our continuous control experiments. We use the same SAC hyperparameters for all methods, other than the fact that we use 3e-4 rather than 3e-5 as the learning rate for vanilla SAC. We use Adam for all optimization. We use three layer networks of width 256 for all function approximation. We use ReLU activations for the actor and critic.

\begin{table}[h]
\begin{center}
\begin{small}
\begin{sc}
\setlength{\tabcolsep}{2pt}
\begin{tabular}{lccccccccccr}
\toprule
 Parameter & Value \\
\midrule
 max replay size & 1000000 \\
 min replay size & 10000 \\
 batch size & 256 \\
 $\gamma$ & 0.99 \\
 $\tau$ & 0.005 \\
 n step & 1 \\
 policy learning rate & 3e-5 \\
 batch size & 256 \\
 samples per insert & 256 \\
 samples per insert tolerance ratio & 0.1 \\
 num SGD steps per step & 1\\
\bottomrule
\end{tabular}
\end{sc}
\end{small}
\end{center}
\caption{\label{table:stbsln3params} Hyperparameters for SAC.}
\end{table}

\paragraph{SPO Parameters.} We use a queue length of $B=10$ and a maximum replay buffer size of $1M$ for non-Markov and intransitive preferences.
For max-reward preferences (excl. and incl. noise), we use a queue length of $B=100$  due to the higher noise in the preferences. We further use a shorter maximum replay buffer size of $100k$ to avoid preference information becoming too stale due to the increased queue length.

\paragraph{Reward Model Parameters.} We use Leaky ReLU activations with a final tanh (following \citet{lee2021pebble}) for the reward model. We update the reward model every 256 policy updates, use a batch size of 64, and a learning rate of 3e-4.

\subsection{Continuous Control Experiments with Intransitive Preferences}
We use the MuJoCo Gym~\citep{Brockman16Gym} Ant-v3 environment as the base environment. The hyper-parameters for SAC are identical to what is described above except for having a fixed entropy coefficient of 1e-4 since the learned policy must exhibit stochasticity to capture the MW; we use a queue size of 100. The preference function we design is composed of a distance and angular preference. The angular preference makes the trajectory lose to an angle of $\theta=\pi/4$ in front of them. The distance component encourages the agent to traverse non-trivial distance from the origin until it hits a certain threshold. We found the distance component to be necessary in order for the agent to maintain effective control of the angle as it is very easy to shift angles when the agent is very close to the origin. The Python code for the preference function is written below.
\begin{python}
def intransitive_ant_preference(traj_1, traj_2):
    return 0.3 * distance_preference(traj_1.radius, traj_2.radius, 10.0) + 
           0.7 * angular_preference(traj_1.angle, traj_2.angle, math.pi/4)

def angular_preference(traj_1_angle, traj_2_angle, angle):
    difference = math.fmod(traj_1_angle + angle/2.0 - traj_2_angle, 2 * math.pi)
    return difference < theta/2.0 or difference > 2 * math.pi - theta/2.0

def distance_preference(traj_1_dist, traj_2_dist, dist_threshold):
    if traj_1_dist > dist_threshold and traj_2_dist > dist_threshold:
        return 1.0
    else:
        return 1.0 if traj_1_dist > traj_2_dist else 0.0

\end{python}

\end{document}